\documentclass{llncs}

\usepackage{subcaption}
\usepackage{graphicx}
\usepackage{setspace}
\usepackage{comment}
\usepackage{longtable}
\usepackage{pgfplots}
\usepackage{algcompatible}
\usepackage{algorithm}
\usepackage{booktabs}
\usepackage{xcolor}
\usepackage{colortbl}
\usepackage{cellspace}
\usepackage{bbm}
\usepackage{booktabs} 
\usepackage[nointegrals]{wasysym}
\usepackage{stmaryrd}
\usepackage{mathtools}
\usepackage{xspace}
\usepackage{array}
\usepackage{makecell}
\usepackage{multirow}
\usepackage{rotating}
\usepackage{soul}
\usepackage{xspace}
\usepackage{nicematrix}
\usepackage{amsmath}
\usepackage{amssymb}
\usepackage{enumitem}
\usepackage{setspace}
\usepackage{float}
\usepackage{tcolorbox}
\usepackage{placeins}

\usepackage{listings}
\makeatletter
\newcommand{\srcsize}{\@setfontsize{\srcsize}{5pt}{5pt}}
\makeatother
\newcommand{\circleempty}{\Circle}
\newcommand{\circlefull}{\CIRCLE}
\newcommand{\circlesemi}{\LEFTcircle}
\definecolor{mygreen}{rgb}{0.2, 0.57, 0.26}

\newcommand{\MIRAGEA}{$\mathtt{MIRAGE}$-$19$\xspace}
\newcommand{\MIRAGEB}{$\mathtt{MIRAGE}$-$22$\xspace}

\newcommand{\HOMECAMPUS}{$\mathtt{Enterprise}$\xspace}

\newcommand{\BESTPERF}{\cellcolor{green!20!white}}
\newcommand{\WORSTPERF}{\cellcolor{red!8!white}}
\newcommand{\CESNETTLS}{$\mathtt{CESNET\text{-}TLS}22$\xspace}
\newcommand{\CESNETQUIC}{$\mathtt{CESNET\text{-}QUIC}23$\xspace}

\newcommand{\tinytiny}[1]{\fontsize{6}{10}\selectfont \textcolor{gray}{$\pm #1$}}

\newcommand{\extrtiny}[1]{\fontsize{2}{4}\selectfont \textcolor{gray}{$\pm #1$}}

\newtcolorbox{detailbox}{
colback=gray!10!white,
colframe=gray!50!white,
fonttitle=\bfseries,
boxrule=0pt,
parbox=false,
boxsep=1pt,
left=2pt,
right=2pt,
top=2pt,
bottom=2pt
}

\begin{document}

\title{Data Augmentation for Traffic Classification}

\author{
    Chao Wang\inst{1,2}
    \and
    Alessandro Finamore\inst{1}
    \and
    Pietro Michiardi\inst{2}
    \and
    \\
    Massimo Gallo\inst{1}
    \and
    Dario Rossi\inst{1}
    \institute{
        Huawei Technologies SASU, France
        \and
        EURECOM, France
    }
}

\maketitle           

\begin{abstract}
Data Augmentation (DA)---enriching training data by adding synthetic 
samples---is a technique widely adopted in Computer Vision (CV) 
and Natural Language Processing (NLP) tasks
to improve models performance. Yet, DA has 
struggled to gain traction in networking contexts, particularly 
in Traffic Classification (TC) tasks. 
In this work, we fulfill this gap 
by benchmarking $18$ augmentation functions 
applied to $3$ TC datasets using packet time series
as input representation and considering a variety of training conditions.
Our results show that ($i$) DA can reap benefits previously unexplored, ($ii$) 
augmentations acting on time series sequence order and masking 
are better suited for TC than amplitude augmentations and 
($iii$) basic models latent space analysis can help understanding
the positive/negative effects of augmentations on classification performance.
\end{abstract}

\section{Introduction}
Network monitoring is at the core of network operations with
Traffic Classification (TC) being key for traffic management.
Traditional Deep Packet Inspection (DPI) techniques, i.e., classifying traffic
with rules related to packets content, is nowadays more and more challenged by the growth in adoption of 
TLS/DNSSEC/HTTPS. Despite the quest for alternative solutions to DPI already sparked about two
decades ago with the first Machine Learning (ML) models based on packet and flow features, 
a renewed thrust in addressing TC via data-driven modeling is fueled today by the rise of Deep Learning (DL),
with abundant TC literature, periodically surveyed~\cite{pacheco2019-tc-survey,shen2023-tc-survey},
reusing/adapting Computer Vision (CV) training algorithms and 
model architectures.

Despite the existing literature, 
we argue that \emph{opportunities laying in the data itself are still unexplored}
based on three observations. First, CV and Natural Language Processing (NLP) methods
usually leverage ``cheap'' Data Augmentation (DA)
strategies (e.g., image rotation or 
synonym replacement) to complement training data by increasing samples variety.
Empirical studies show that this leads to improved classification accuracy.
Yet to the best of our knowledge, only a handful of TC studies considered 
DA~\cite{horowicz2022miniflowpic,rezaei2019ICDM-ucdavis,Xie2023DACT4TC}
and multiple aspects of DA design space remain unexplored.
Second, network traffic datasets are imbalanced
due to the natural skew of app/service popularity and traffic dynamics.
In turn, this calls for training strategies emphasizing classification 
performance improvement for classes with fewer samples.
However, the interplay between imbalance and model performance
is typically ignored in TC literature. 
Last, the pursuit of better model generalization and robustness necessitates 
large-scale datasets with high-quality labeling resulting in expensive data 
collection processes. In this context, the extent to which DA can alleviate 
this burden remains unexplored. 

In this paper, we fill these gaps by providing
a comprehensive evaluation of ``hand-crafted'' augmentations---transformations
designed based on domain knowledge---applied
to packets time series typically used as input in TC.
Given the broad design space, we
defined research goals across
multiple dimensions.
First of all, we selected a large pool of 18 augmentations
across 3 families (amplitude, masking, and sequence)
which we benchmark both when used in isolation as well
as when multiple augmentations are combined (e.g., via stacking or ensembling).
Augmentations are combined with original training data
via different batching policies (e.g., replacing training data with augmentation,
adding augmented data to each training step, or 
pre-augmenting the dataset before training).
We also
included scenarios where imbalanced 
datasets are re-balanced during training 
to give more importance to minority classes.
Last, we dissected augmentations performance by
exploring their geometry in the classifiers
latent space to pinpoint root causes driving performance. 
Our experimental campaigns were carried over 2 mid-sized public datasets,
namely \MIRAGEA and \MIRAGEB (up to 20 classes, 64k flows), 
and a larger private dataset (100 classes, 2.9M flows).
We summarize our major findings as follows:

\begin{itemize}

\item We confirm that augmentations 
improve performance (up to +4.4\% 
weighted F1) and expanding
training batches during training (i.e., the Injection policy)
is the most effective policy to introduce augmentations.
Yet, improvements are dataset dependent 
and not necessarily linearly related to dataset size
or number of classes to model;

\vspace{5pt}
\item Sequence ordering and masking are
more effective augmentation families
for TC tasks. Yet, no single augmentation
is found consistently superior across datasets,
nor domain knowledge suffice
to craft effective augmentations, i.e.,
the quest for effective augmentations is an
intrinsic trial-and-error process;

\vspace{5pt}
\item Effective augmentations introduce
good sample variety, i.e., they synthesize
samples that are neither too close nor
too far from the original training data.

\end{itemize}

To the best of our knowledge, a broad and systematic study 
of hand-crafted DA techniques in TC as the one
performed in our study is unprecedented.
Ultimately, our analysis confirms that DA
is currently suffering from a single pain point---exploring
the design space via brute force.
However, our results
suggest a possible road map to achieve better augmentations
via generative models which might 
render obsolete the use of brute force.

In the remainder, we start by introducing DA
basic concepts and reviewing
relevant ML and TC literature (Sec.~\ref{sec:related}).
We then introduce and discuss our research goals (Sec.~\ref{sec:goals})
and the experimental setting used to address them (Sec.~\ref{sec:experimental_settings}).
Last, we present our results (Sec.~\ref{sec:results}) before 
closing with final remarks (Sec.~\ref{sec:conclusions}).

\section{Background and related work}
\label{sec:related}

Data augmentation consists in adding synthetic samples (typically derived from real ones) 
to the training set to increase its variety. 
DA has been popularized across many ML disciplines~\cite{MUMUNI2022DAsurvey,Shorten2019imgeDAsurvey,Wen2021TS_aug_survey}
with a large number of variants which we can be broadly grouped into
two categories~\cite{MUMUNI2022DAsurvey}: hand-crafted DA and data synthesis.
Hand-crafted DA (also known as data transformations) 
involves creating new samples by applying predefined rules to existing samples. 
Instead, data synthesis relates to generating new samples 
via generative models, e.g., Variational AutoEncoders (VAE), 
Generative Adversarial Neural networks (GAN), Diffusion Models (DM), etc., trained on existing and typically large datasets.

In this section, we overview the existing DA
literature with an emphasis
on hand-crafted DA and methods closer to the scope of our work. 
We begin by introducing relevant CV and time series ML literature.
Then, we review TC literature using DA and close with a discussion
about general design principles/requirements that we used for 
defining our research goals outlined in Sec.~\ref{sec:goals}.

\subsection{Data augmentation in traditional machine learning tasks}

To ground the discussion of different methods merits, we start by
revisiting the internal mechanisms of supervised ML/DL models.

\vspace{3pt}
\noindent \textbf{Supervised modeling and DA.}
In a nutshell, a supervised model is a function $\varphi: {\bf x}\in\mathcal{X} \rightarrow y\in\mathcal{Y}$ mapping an input
$\bf x$ to its label $y$. Training such models corresponds to 
discover a good function $\varphi(\cdot)$ based on a training set.
When performing DA, the training set is enlarged by adding 
new samples ${\bf x'} = \textrm{Aug}({\bf x})$ created by altering original samples ${\bf x}$---these 
transformations act directly in the input space $\mathcal{X}$ and the additional
synthetic samples contribute in defining $\varphi(\cdot)$ as much as the original
ones. It follows that having a comprehensive understanding of
samples/classes properties and their contribution
to models training is beneficial for 
designing \emph{effective} augmentations, i.e.,
transformations enabling higher classification performance.

Beside operating in the input space, DL models offer also a
latent space. In fact, DL models are typically a composition of two functions $\varphi({\bf x_i}) = h(f({\bf x_i})) = y_i$:
a feature extractor $f(\cdot)$ and a classifier $h(\cdot)$, normally a single
fully connected layer (i.e., a linear classifier) in TC.
In other words, an input sample $f({\bf x_i})={\bf z_i}$ is first projected
into an intermediate space, namely the \emph{latent space}, where
different classes are expected to occupy different regions.
The better such separation, the easier is for the classifier $h({\bf z_i}) = y_i$
to identify the correct label.
It follows that this design enables a second form of augmentations based
on altering samples in the latent space rather than in the input space.

Last, differently from DA, generative models aims to learn the
training set data distribution. In this way generating 
new synthetic data corresponds to sampling from the learned distribution.
In the following we expand on each of these three methodologies.

\paragraph{\bf Input space transformations.} 
In traditional ML, Synthetic Minority Over-sampling 
TEchnique (SMOTE)~\cite{Chawla2002smote} is a popular 
augmentation technique. This approach 
generates new samples by interpolating the nearest 
neighbors of a given training sample. To address class 
imbalance, SMOTE is often employed with a sampling 
mechanism that prioritizes minority 
classes~\cite{Han2005BorderlineSMOTE,He2008ADASYN}.

In CV, several image transformations have been 
proposed to improve samples variety while 
preserving classes semantics. These transformations operate on colors 
(e.g., contrast and brightness changes, gray scaling)
and geometry (e.g., rotation, flipping, and zooming), 
or via filters (e.g., blurring with Gaussian kernel)
and masks (e.g., randomly set to zero a patch of pixels).
Furthermore, transformations like CutMix~\cite{yun2019cutmix} 
and Mixup~\cite{zhang2018mixup} not only increase 
samples variety but also increase \emph{classes variety} by creating 
synthetic classes from a linear combination of existing ones. 
The rationale behind this approach is that by introducing new artificial 
classes sharing similarities with the 
true classes the classification task becomes intentionally 
more complex, thereby pushing the training process to 
extract better data representations.
Empirical validations of DA techniques in CV have consistently 
demonstrated their effectiveness across a diverse range of datasets, 
tasks, and training paradigms~\cite{chen2020simclr,he2015resnet,redmon2016yolo}. 
As a result, DA has become a ubiquitous component in the CV models training pipelines.

Considering time series instead, input transformations can
either modify data \emph{amplitude} (e.g., additive 
Gaussian noise) or manipulate \emph{time} (e.g., composing 
new time series by combining different segments of existing ones).
Similarly to CV, the research community has provided 
empirical evidence supporting the effectiveness of these 
transformations in biobehavioral~\cite{yang2022bioDA} and 
health~\cite{yu2022stressDAsemi} domains.
However, contrarily to CV, these transformations are less diverse 
and have been less widely adopted, possibly due to the stronger
reliance on domain knowledge---an amplitude change 
on an electrocardiogram can be more difficult
to properly tune compared to simply rotating an image.

\paragraph{\bf Latent space transformations.}
Differently from traditional ML, DL models offer the ability
to shape the feature extractor to create
more ``abstract'' features. 
For example, Implicit Semantic Data Augmentation (ISDA)~\cite{wang2020ISDA} first 
computes class-conditional covariance matrices based on intra-class feature variety;
then, it augments features by translating real features along random directions 
sampled from a Gaussian distribution defined by the class-conditional covariance matrix. 
To avoid computational inefficiencies caused by explicitly augmenting each sample 
many times, 
ISDA computes an upper bound of the expected cross entropy loss on an enlarged 
feature set and takes this upper bound as the new loss function.
Based on ISDA, and focused on data imbalance, Sample-Adaptive Feature Augmentation 
(SAFA)~\cite{Hong2022SAFA} extracts transferable features from the majority classes 
and translates features from the minority classes in accordance with the 
extracted semantic directions for augmentation.

\paragraph{\bf Generative models.}
In addition to traditional hand-crafted data augmentation techniques, generative models 
offer an alternative solution to generate samples variety.
For instance, \cite{burg2023DApersdiffusion,dafusion}
use a multi-modal diffusion model trained on an
Internet-scale dataset composed of (image, text) pairs.
Then, the model is used to synthesize new 
samples---text prompts tailored to specific 
downstream classification tasks are used as conditioning
signal to create task-specific samples---to 
enlarge the training set for a classification task.
While these types of generative models can provide high-quality
samples variety, their design and application still requires
a considerable amount of domain knowledge to be effective.

\subsection{Data augmentation in traffic classification \label{sec:related-da-tc}}

TC tasks usually rely on either packet time series
(e.g., packet size, direction, Inter Arrival Time (IAT), 
etc., of the first 10-30 packets of a flow) or payload bytes (e.g.,
the first 784 bytes of a flow, possibly gathered by concatenating
payload across different packets) arranged as 2d matrices.
Recent literature also considers combining both input types into multi-modal
architectures~\cite{aceto2019mimetic,akbari2022sigmetrics,cesnetquicTMA23}.

Such input representations and datasets exhibit three notable distinctions when compared to data from other ML/DL disciplines. 
First, TC datasets show \emph{significant class imbalance}---this is
a ``native'' property of network traffic as different applications
enjoy different popularity and traffic dynamics while, for instance,
many CV datasets are balanced. Second, TC input representation is typically \emph{``small''} 
to adhere to desirable system design properties---network
traffic should be ($i$) \emph{early classified}, i.e., the application associated
to a flow should be identified within the first few packets of a flow, 
and ($ii$) computational/memory resources required to represent a flow should be
minimal as an in-network TC systems need to cope with hundreds of thousands 
of flow per second. Last, TC input data has \emph{weak semantics}---the 
underlying application protocols (which may or not be known a priori) may not be 
easy to interpret even for domain experts when visually inspecting packet time series.

\paragraph{\bf Hand-crafted DA.}
The combination of the above observations leads to have only a
handful of studies adopting DA in TC.
Rezaei et al.~\cite{rezaei2019ICDM-ucdavis} created synthetic input samples 
by means of three hand-crafted DA strategies based on sampling multiple short sequences across the duration of a complete flow.
Horowicz et al.~\cite{horowicz2022miniflowpic} instead focused on a \textit{flowpic} input representation---a 2d summary of the evolution of packets size throughout the duration of a flow---augmented by first altering the time series collected from the first 15s of a flow and used to compute the flowpic. 
While both studies show the benefit of DA,
these strategies violate the early classification principle as
they both consider multiple seconds of traffic, thus they are
better suited for post-mortem analysis only.
Conversely, Xie et al.~\cite{Xie2023DACT4TC} recently proposed
some packet series hand-crafted DA to tackle
data shifts arising when applying a model on network traffic
gathered from networks different from the ones used to collect
the training dataset. Specifically, inspired by TCP protocol 
dynamics, authors proposed five packets time series augmentations (e.g., to mimic a packet lost/retransmission one can replicate a value at a later position in the time series)
showing that they help to mitigate data shifts.
Yet, differently from~\cite{horowicz2022miniflowpic,rezaei2019ICDM-ucdavis}, 
the study in~\cite{Xie2023DACT4TC} lacks from an ablation of each individual augmentation's performance.

\paragraph{\bf Generative models.}
Last, \cite{Wang2020PacketCGAN,Wang2019FLOWGAN,Yin2018BotGAN} 
investigate augmentations based on GAN methods when using payload bytes as input for 
intrusion detection scenarios, i.e., 
a very special case of TC where the classification task is binary.
More recently, \cite{xi2023hotnets-generative-traces} compared
GAN and diffusion model for generating raw payload bytes traces
while \cite{sivaroopan23synig,sivaroopan2023netdiffus} instead
leveraged GAN or diffusion models to generate 
2D representations (namely GASF) of longer traffic flow signals 
for downstream traffic
fingerprinting, anomaly detection, and TC.

\subsection{Design space}

\paragraph{\bf Search space.} 
Independently from the methodology and application discipline,
DA performance can only be assessed via empirical studies,
i.e., results are bound to the scenarios and the datasets used.
Moreover, to find an efficient strategy one should
consider an array of options, each likely
subject to a different parametrization.
In the case of hand-crafted DA, one can also
opt for using \emph{stacking} (i.e., applying a sequence of transformations) or
\emph{ensembling} (i.e., applying augmentations by selecting
from a pool of candidates according to some sampling logic)---an 
exhaustive grid search is unfeasible given the large search space.
Besides following guidelines to reduce the number of options~\cite{cubuk2019randaugment},
some studies suggest the use of reinforcement learning to guide the search space exploration~\cite{cubuk2019autoaugment}. 
Yet, no standard practice has emerged.

\paragraph{\bf Quantifying good variety.}
As observed in TC literature~\cite{horowicz2022miniflowpic,rezaei2019ICDM-ucdavis,Xie2023DACT4TC}, domain knowledge 
is key to design efficient augmentations. Yet, ingenuity might not be enough
as models are commonly used as ``black boxes'', making it extremely challenging to establish a direct link between an augmentation technique and its impact on the final classification performance.  
For instance,
rotation is considered a good image transformation as result of
empirical studies. Likewise, generative models are trained
on large image datasets but without an explicit connection
to a classification task~\cite{schuhmann2022laionb}---the design of the 
augmentation method itself is part of a trial-and-error approach
and the definition of metrics quantifying the augmentation quality
is still an open question.

One of the aspects to be considered when formulating such metrics is the \emph{variety}
introduced by the augmentations.
Gontijo-Lopes et al.~\cite{Cubuk2021Tradeoff} propose metrics quantifying the distribution shift and diversity introduced by DA
contrasting models performance with and without augmentations.
Other literature instead focuses on mechanisms that
can help defining desirable properties for augmentations.
For instance, from the feature learning literature, \cite{shen2022FeatureManipu,zou2023MixupRareF} find that 
DA induces models to better learn
rare/less popular but good features by altering their importance, thus improving model generalization performance.
Samyak et al.~\cite{Jain2023Aggregate} find that optimization trajectories are different when training on different augmentations and propose to aggregate the weights of models trained on different augmentations to obtain a more uniform distribution of feature patches, encouraging the learning of diverse and robust features.

\paragraph{\bf Training loss.} 
Self-supervision and contrastive learning are DL
training strategies that take advantage of augmentations by design.
In a nutshell, contrastive learning consists of a 2-steps training process.
First, a feature extractor is trained in a \emph{self-supervised} manner with a contrastive loss function that
pulls together different augmented ``views'' of a given sample 
while distancing them from views of other samples. Then, a classifier head is trained on top of
the learned representation 
in a \emph{supervised} manner
using a few labeled samples---the better the feature representation, the
lower the number of labeled samples required for 
training the head.
Empirical studies have demonstrated the robustness of the 
feature representations learned with contrastive learning~\cite{chen2020simclr}
and a few recent studies investigated contrastive learning also in TC~\cite{guarino2023tma,horowicz2022miniflowpic,towhid-netsoft22-boyl,Xie2023DACT4TC}.

\paragraph{\bf Linking generative models to classifiers.} 
When we consider the specific case of using generative models to
augment training data, we face a major challenge---generative
models are not designed to target a specific downstream task~\cite{sivaroopan23synig,sivaroopan2023netdiffus,Wang2019FLOWGAN}.
While studies like~\cite{odena2017acgan} integrated a classifier in GAN training in the pursuit of improving the reliability the model, how to properly link
and train a generative model to be sensitive to a downstream 
classification task is still an open question
even in CV literature.

\section{Our goals and methodology}
\label{sec:goals}
Drawing insights from the literature reviewed in Sec.~\ref{sec:related},  
we undertake a set of empirical campaigns to better understand hand-crafted DA when applied 
in the input space for TC task and address the following research goals:

\begin{enumerate}[label={\bf {G\arabic*}.},start=1,leftmargin=2\parindent]
    \item How to compare the performance of different augmentations? 
    This includes investigating augmentations sensitivity to their 
    hyper-parametrization and dataset properties (e.g., number of samples and classes).
    
    \vspace{8pt}
    \item How augmented samples should be added to the training set and
    how many samples should be added? Is augmenting minority classes beneficial
    to mitigate class imbalance?

    \vspace{8pt}
    \item Why some augmentations are more effective than others?

    \vspace{8pt}
    \item Does combining multiple augmentations provide extra performance improvement?

\end{enumerate}

In the remainder of this section, we motivate each goal and 
introduce the methodology we adopted to address them.

\begin{figure}[t]
\centering
\includegraphics[width=\textwidth]{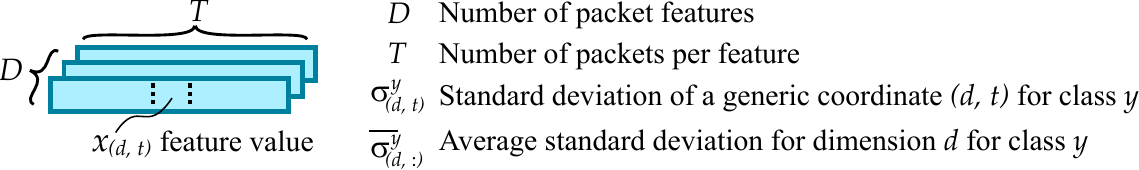}
\caption{Input sample ${\bf x}$ shape and related notation.
\label{fig:input-sample}
}
\end{figure}

\begingroup
\vspace{-10em}
\begin{table}[h]
\centering
\caption{Amplitude augmentations.}
\label{tab:augs_amplitude}
\setstretch{0.5}
\fontsize{4}{6}\selectfont
\begin{tabular}{
    @{}
    m{0.10\textwidth}
    @{$\,\,$}c
    @{$\,$}c
    @{$\,$}c
    @{}m{0.6\textwidth}
    @{$\,\,$}p{0.22\textwidth}
    @{}
}
\toprule
\multirow{2}{*}{ \bf Name} & 
    \multicolumn{3}{@{}c@{}}{\bf Pkts Feat.} & 
    \multicolumn{1}{@{}c@{}}{\multirow{2}{*}{\bf Description}} & 
    \multicolumn{1}{@{}c@{}}{\bf Example} 
\\
& Size &  Dir & IAT & &
\multicolumn{1}{c}{magnitude $\alpha\!=\!0.5$} 
\\
\midrule
Gaussian Noise & \circlesemi & \circleempty & \circlesemi
&
Add independently sampled Gaussian noise to Size or IAT
\begin{detailbox}
\emph{Details:}
Sample a feature $d\!\!\in\!\!\{\rm Size, IAT\}$ and
add Gaussian noise to its values $x_{(d,t)}\!+\!\varepsilon_t$ 
where $\varepsilon_t\!\!\sim\!\!\mathcal{N}(0, \alpha \{{\sigma_{(d,t)}^y}\}^2)$
\end{detailbox}
&      
\begin{minipage}{0.22\textwidth}
\includegraphics[width=\linewidth]{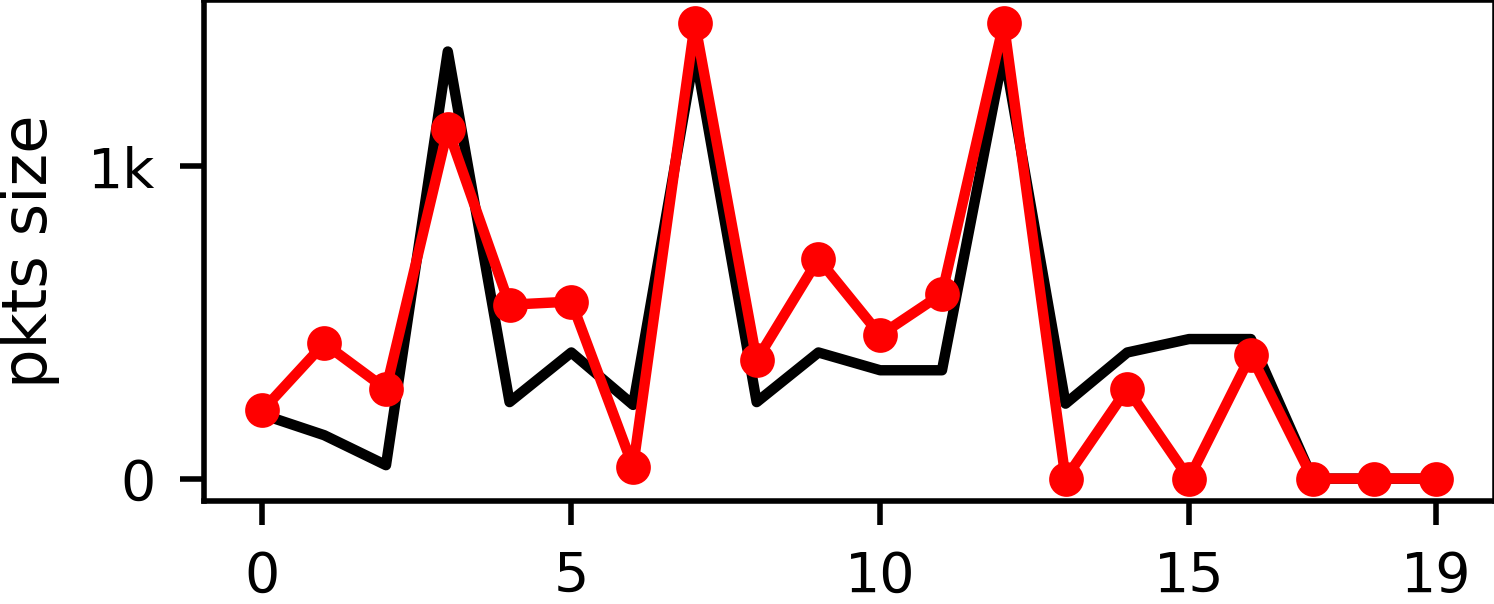}
\end{minipage}
\\
\noalign{\global\arrayrulewidth=1px}
\arrayrulecolor{gray!10}
\hline
Spike Noise$\,$\cite{Wen2021TS_aug_survey}  & \circlesemi & \circleempty & \circlesemi
&  
Add independently sampled Gaussian noise to Size or IAT
\begin{detailbox}
\emph{Details:}
Sample a feature $d\!\!\in\!\!\{\rm Size, IAT\}$
and add Gaussian noise to up to 3 of its non-zero values 
$x_{(d,t)}\!+\!|\varepsilon_t|$ where $\varepsilon_t\!\!\sim\!\!\mathcal{N}(0, \alpha \{{\sigma_{(d,t)}^y}\}^2)$ 
\end{detailbox}
&      
\begin{minipage}{0.22\textwidth}
\includegraphics[width=\linewidth]{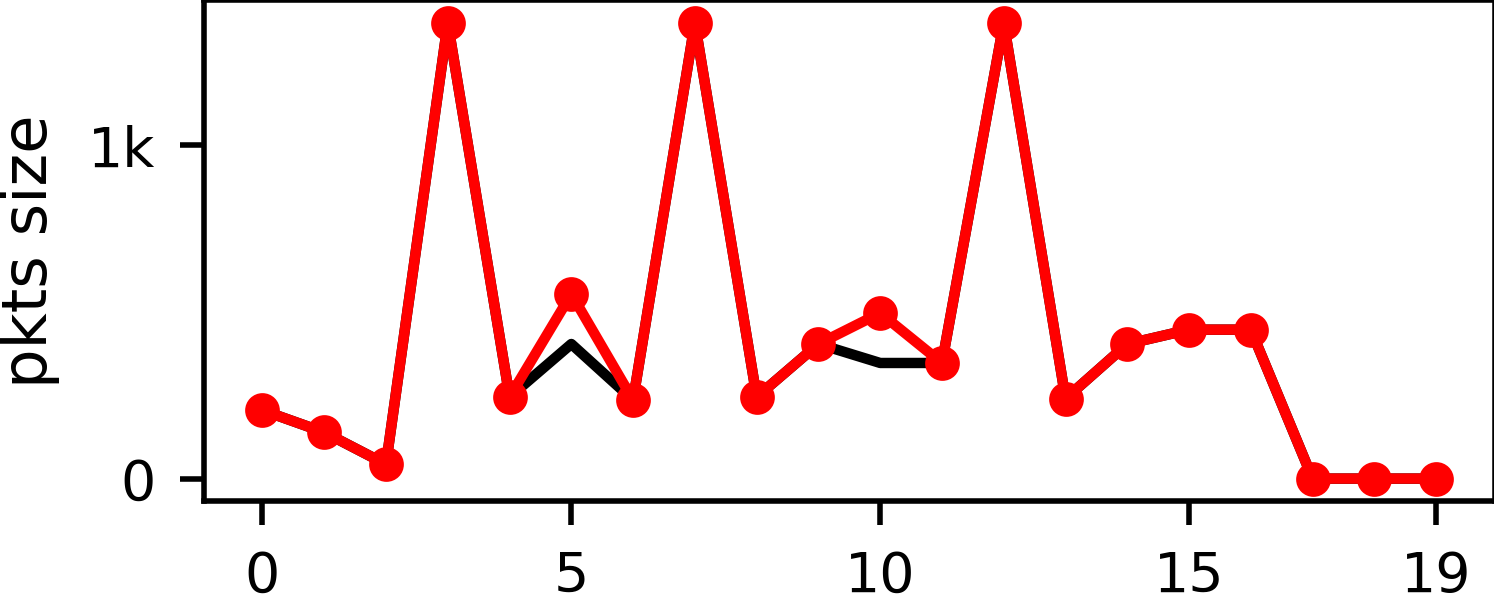}
\end{minipage}
\\
\hline
Gaussian WrapUp$\,$\cite{Eldele2021ts_strong_weak_ct}  & \circlesemi & \circleempty & \circlesemi
&  
Scale Size or IAT by independently sampled Gaussian values
\begin{detailbox}
\emph{Details:}
Sample a feature $d\!\!\in\!\!\{\rm Size, IAT\}$
and multiply Gaussian noise to its values $x_{(d,t)} \cdot \varepsilon_t$ with \newline
$\epsilon_t\!\!\sim\!\!\mathcal{N}(1+0.01\alpha, 0.02\alpha \{{\sigma_{(d,t)}^y}\}^2)$
\end{detailbox}
&      
\begin{minipage}{0.22\textwidth}
\includegraphics[width=\linewidth]{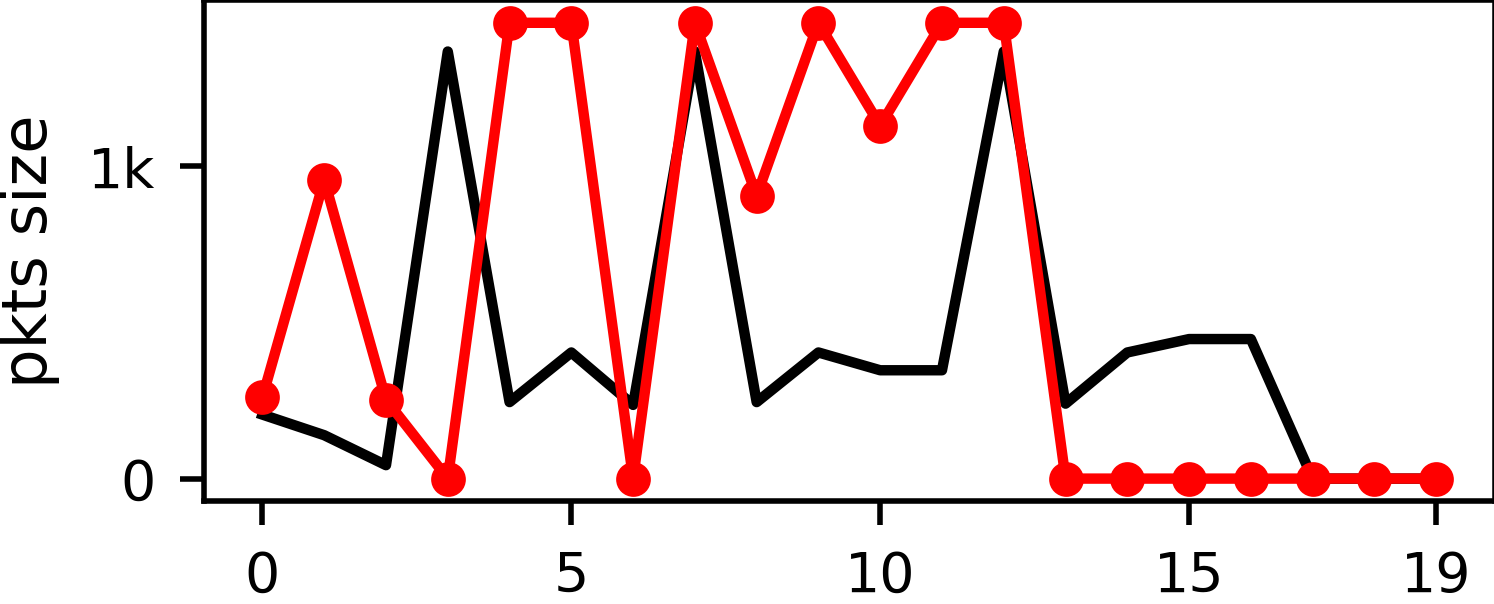}
\end{minipage}
\\
\hline
Sine WrapUp~\cite{Johannes2022CT_TS_Applied_Soft_Computing}  & \circlesemi & \circleempty & \circlesemi
&  
Scale Size or IAT by sinusoidal noise
\begin{detailbox}
\emph{Details:}
Sample a feature $d\!\!\in\!\!\{\rm Size, IAT\}$
and multiply its values by a sine-like
noise $x_{(d,t)}\cdot\epsilon_i$ with 
\mbox{$\epsilon_i\!=\![1+0.02\alpha \cdot \overline{\sigma_{(d, :)}^y} \cdot \sin(\frac{4\pi i}{T} + \theta)]$}
and $\theta\!\!\sim\!\!U[0, 2\pi[$
\end{detailbox}
&      
\begin{minipage}{0.22\textwidth}
\includegraphics[width=\linewidth]{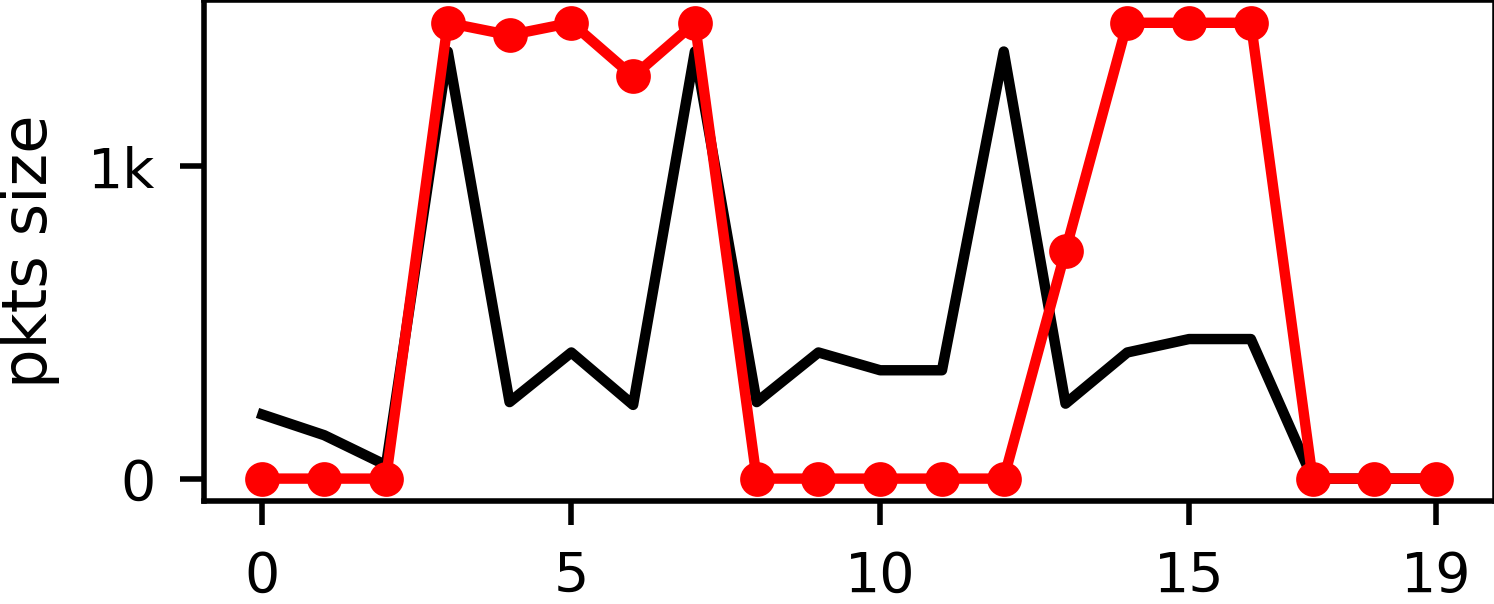}
\end{minipage}
\\
\hline
Constant WrapUp$\,$\cite{horowicz2022miniflowpic}  & \circleempty & \circleempty & \circlefull
&  
Scale IAT by a single randomly sampled value
\begin{detailbox}
\emph{Details:}
Sample a single uniformly sampled value 
\mbox{$\epsilon\!\!\sim\!\!U[a, b]$}
and perform $x_i\cdot\epsilon$ to all
$x_i$ of IAT with
\newline
\mbox{$a = 1+\overline{\sigma_{(d, :)}^y}\cdot(0.06-0.02\alpha)$}; 
\mbox{$b = 1+\overline{\sigma_{(d, :)}^y}\cdot(0.14+0.02\alpha)$} 
\end{detailbox}
&      
\begin{minipage}{0.22\textwidth}
\includegraphics[width=\linewidth]{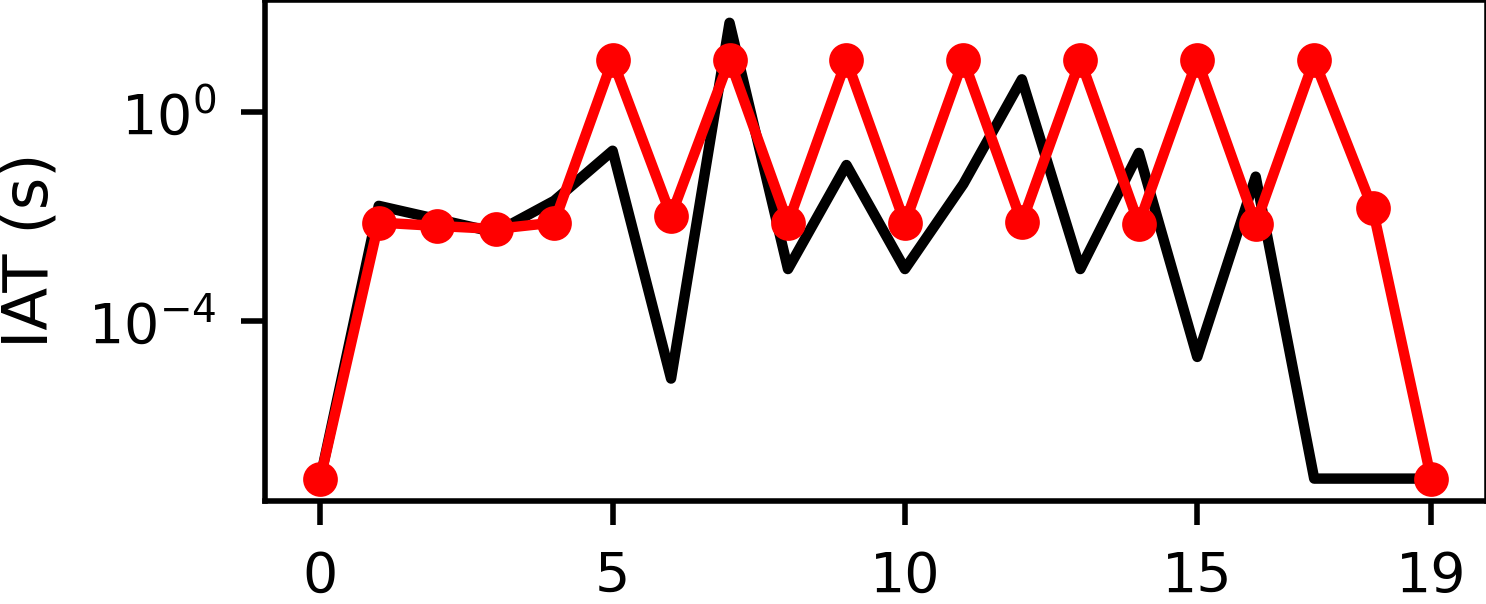}
\end{minipage}
\\
\arrayrulecolor{black}
\bottomrule
\end{tabular}
\\
\raggedright
\circleempty~feature never used;  \circlesemi~feature selected randomly; \circlefull~feature always used.\\
In the figures, black solid lines \rule[0.1em]{1em}{2pt} for original samples, \textcolor{red}{red $\CIRCLE$ lines} for augmented samples; x-axis for time series index and y-axis the feature value (either packet size or IAT).
\end{table}
\endgroup

\begingroup
\vspace{-20em}
\begin{table}[h]
\centering
\caption{Masking augmentations}
\label{tab:augs_mask}
\setstretch{0.7}
\fontsize{4}{5}\selectfont
\vspace{3pt}
\raggedright
All three features (Size, DIR and IAT) are affected by all transformations.
\begin{tabular}[t]{
    @{}
    m{0.10\textwidth}
    @{}m{0.7\textwidth}
    @{$\,\,$}p{0.24\textwidth}
}
\toprule
\multirow{2}{*}{ \bf Name} & 
    \multicolumn{1}{@{}c@{}}{\multirow{2}{*}{\bf Description}} & 
    \multicolumn{1}{@{}c@{}}{\bf Example} 
\\
& &
\multicolumn{1}{c}{magnitude $\alpha\!=\!0.5$} 
\\
\midrule
Bernoulli Mask~\cite{Yue2022TS2Vec}  & 
Random masking values
\begin{detailbox}
\emph{Details:} Independently 
set to zero feature values by sampling a \newline
${\rm Bernoulli}(p = 0.6\alpha)$
\end{detailbox}
&      
\begin{minipage}{0.24\textwidth}
\includegraphics[width=\linewidth]{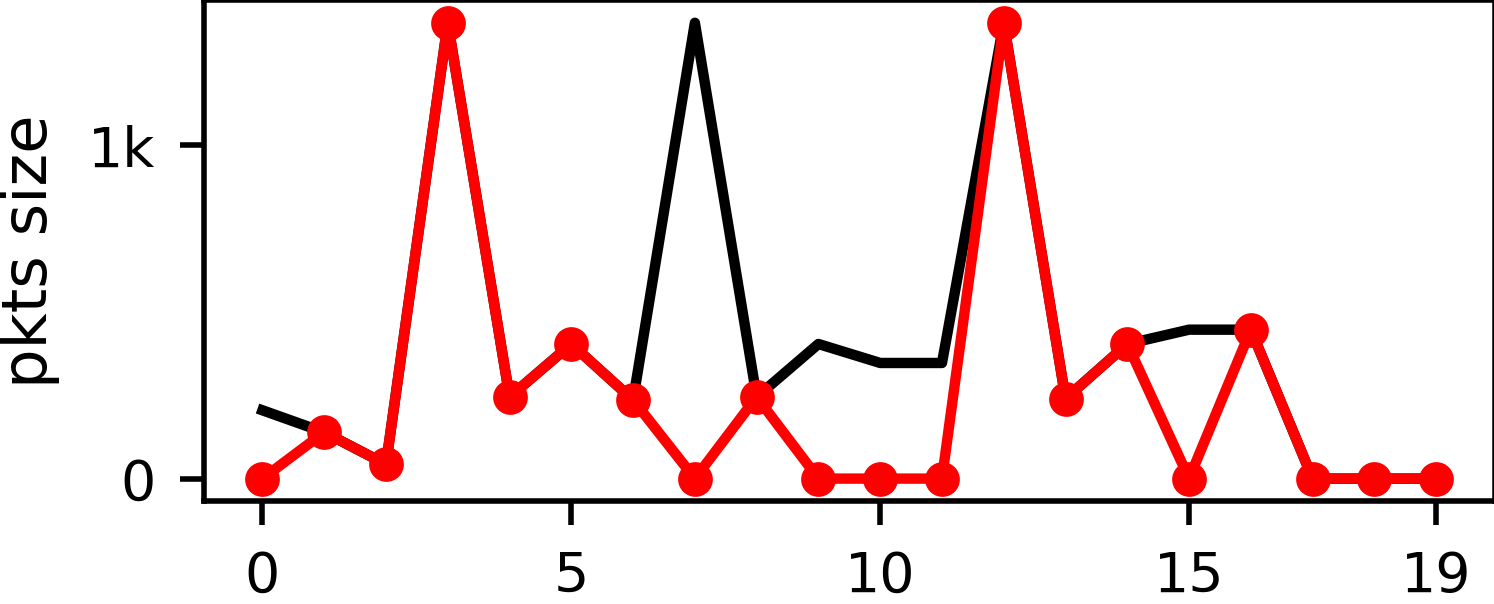}
\end{minipage}
\\
\noalign{\global\arrayrulewidth=1px}
\arrayrulecolor{gray!10}
\hline
Window Mask~\cite{Johannes2022CT_TS_Applied_Soft_Computing}  &
Masking the same sequences across all features
\begin{detailbox}
\emph{Details:}
Given a configured maximum size \mbox{$W\!\!=\!\!\left\lfloor 1\!+\!2.5\alpha \right\rceil$},
sample a window length $w\!\!\sim\!\!U[1, W]$ and a random starting point $t\!\!=\!\!U[0,T-w]$
and set to zero all $x_{(:, t)}$ falling in the sampled window
\end{detailbox}
&      
\begin{minipage}{0.24\textwidth}
\hspace{2em}$w = 2, t = 14$\\
\includegraphics[width=\linewidth]{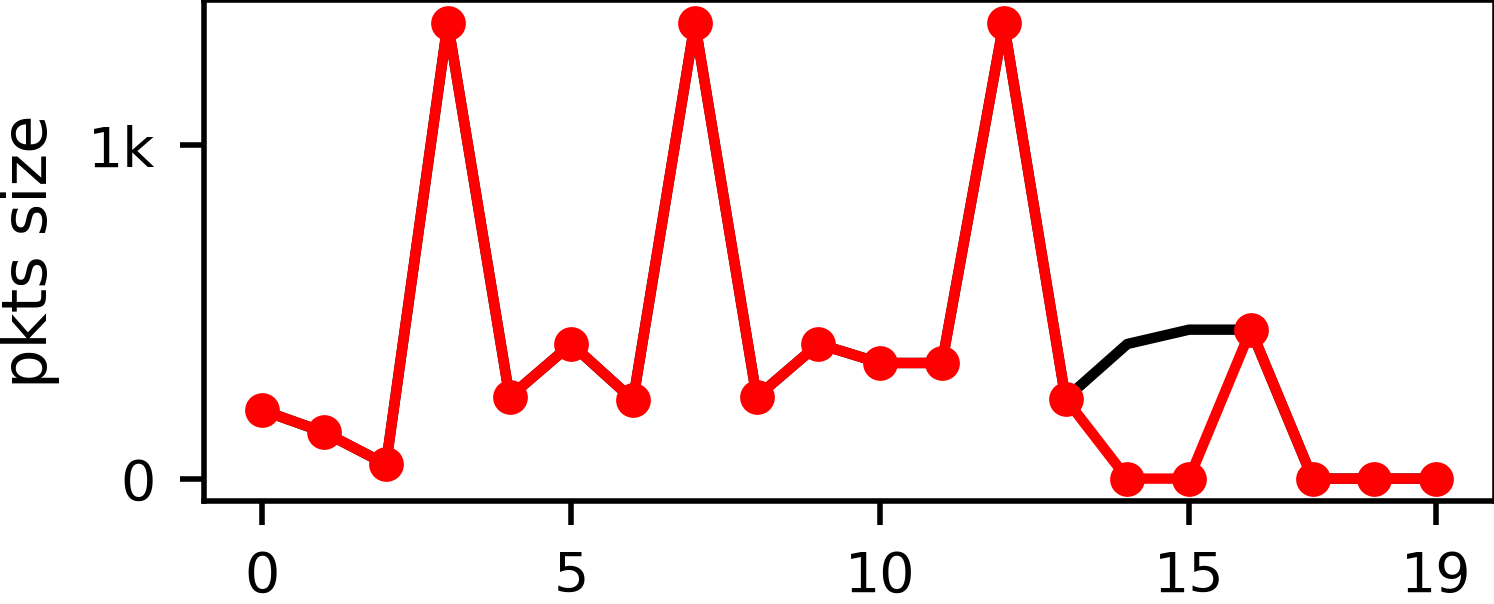}
\end{minipage}
\\
\arrayrulecolor{black}
\bottomrule
\end{tabular}
\\
\raggedright
In the figures, black solid lines \rule[0.1em]{1em}{2pt} for original samples, \textcolor{red}{red $\CIRCLE$ lines} for augmented samples.
\end{table}
\endgroup
\FloatBarrier

\begingroup

\vspace{-5em}
\begin{center}
\setstretch{0.7}
\fontsize{4}{6}\selectfont
\begin{longtable}[h]{
    @{}
    m{0.16\textwidth}
    @{$\,\,$}m{0.65\textwidth}
    @{$\,\,$}p{0.22\textwidth}
}
\caption{Sequence order augmentations.
\label{tab:augs_order}
}
\\
\toprule
\multirow{2}{*}{ \bf Name} & 
    \multicolumn{1}{@{}c@{}}{\multirow{2}{*}{\bf Description}} & 
    \multicolumn{1}{@{}c@{}}{\bf Example}
\\
& &
\multicolumn{1}{c}{magnitude $\alpha\!=\!0.5$} 
\\
\midrule
\endfirsthead
\midrule 
\multicolumn{3}{r}{\textit{Continued on next page}} \\
\endfoot
\hline
\endlastfoot
Horizontal Flip~\cite{Johannes2022CT_TS_Applied_Soft_Computing} & 
Swap values left to right (no magnitude needed)
&      
\begin{minipage}{0.22\textwidth}
\includegraphics[width=\linewidth]{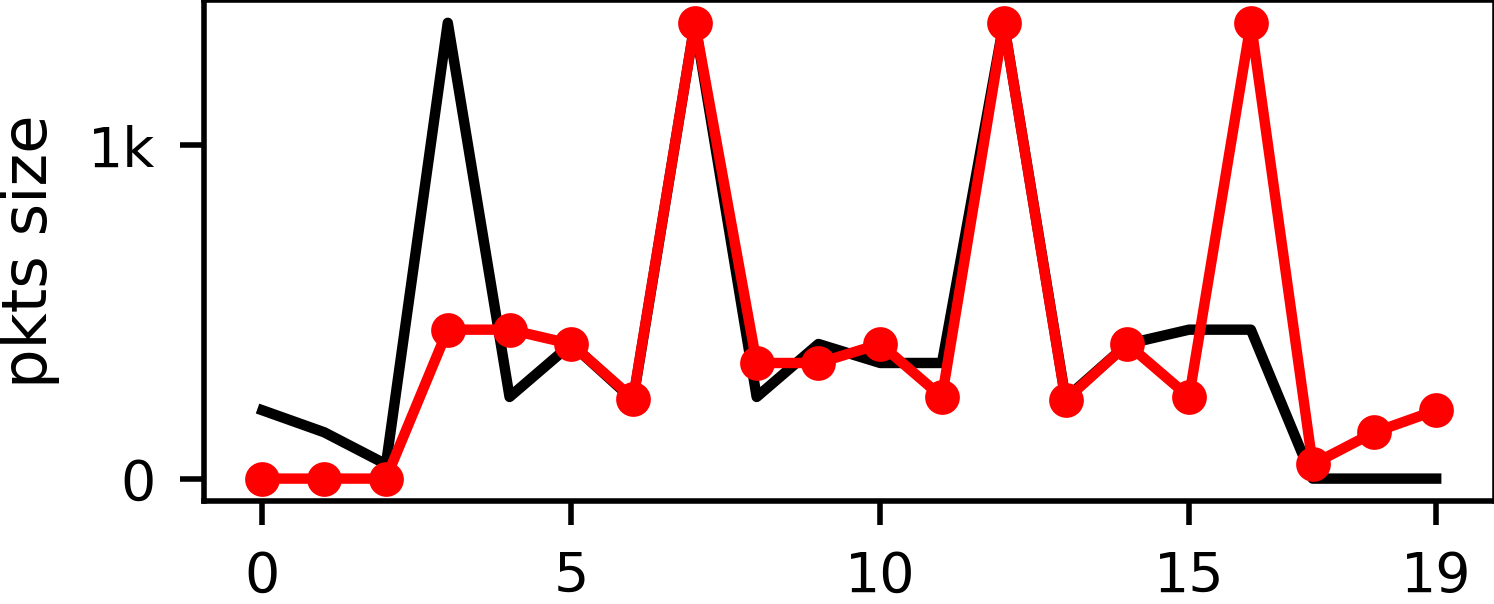}
\end{minipage}
\\
\noalign{\global\arrayrulewidth=1px}
\arrayrulecolor{gray!10}
\hline
Interpolation~\cite{Johannes2022CT_TS_Applied_Soft_Computing}  &    
Densify time series by injecting average values and then sample
a new sequence of length T
\begin{detailbox}
\emph{Details:}
Expand each feature by inserting the average $0.5(x_{(d,t)} +x_{(d,t+1)})$ 
in-between each pair of values. Then
randomly select a starting point $t\!\!=\!\!U[0,T-1]$ 
and extract the following $T$ values for all features 
$x_{(:, t:t+T)}$ (no magnitude needed).
\end{detailbox}
&      
\begin{minipage}{0.22\textwidth}
\hspace{3em}$t = 8$\\
\includegraphics[width=\linewidth]{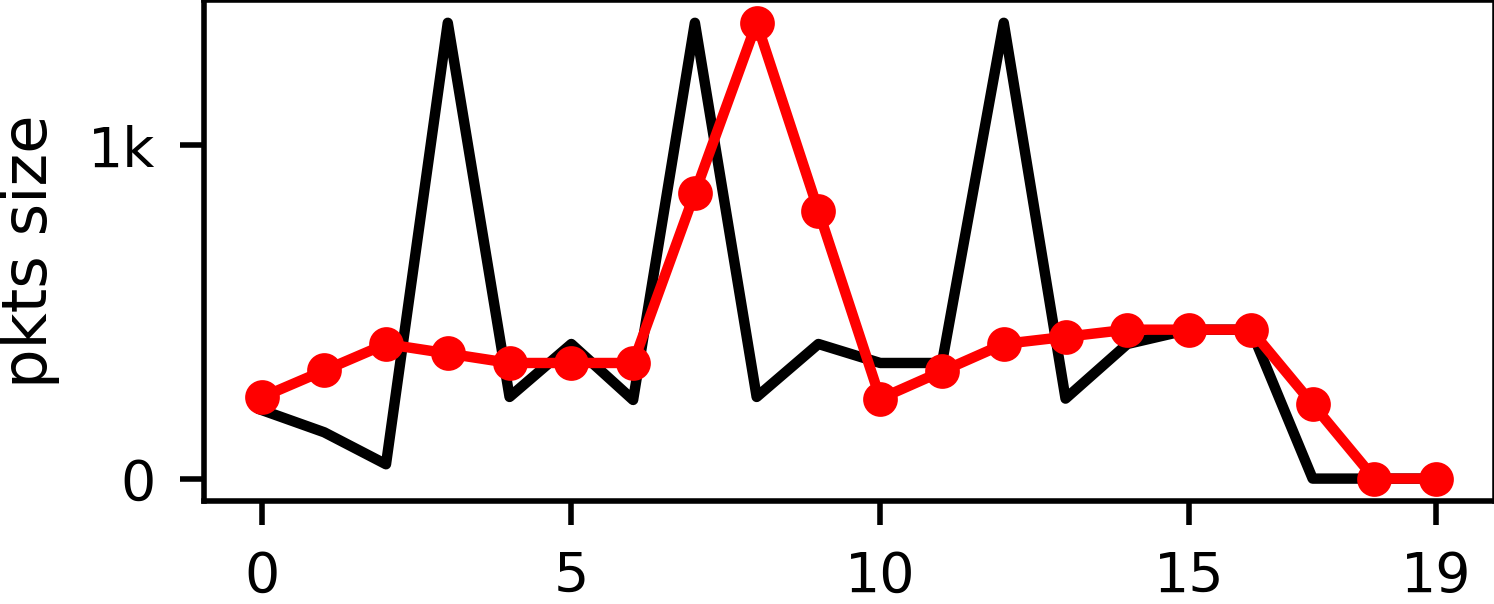}
\end{minipage}
\\
\hline
CutMix~\cite{yun2019cutmix} &
Swap segments of two different samples
\begin{detailbox}
\emph{Details:}
Given a training mini-batch, define pairs of samples (${\bf x1}, {\bf x2})$
by sampling without replacement.
Then sample a \emph{segment}
of length \mbox{$w\!\!\sim\!\!U[0,T-1]$} starting at
\mbox{$t\!\!\sim\!\!U[0,T-1-w]$} and swap
the segment of each feature between ${\bf x1}$ and ${\bf x2}$
(no magnitude needed).
\end{detailbox}
&      
\begin{minipage}{0.22\textwidth}
\centering $w = 8, t = 2$\\
\includegraphics[width=\linewidth]{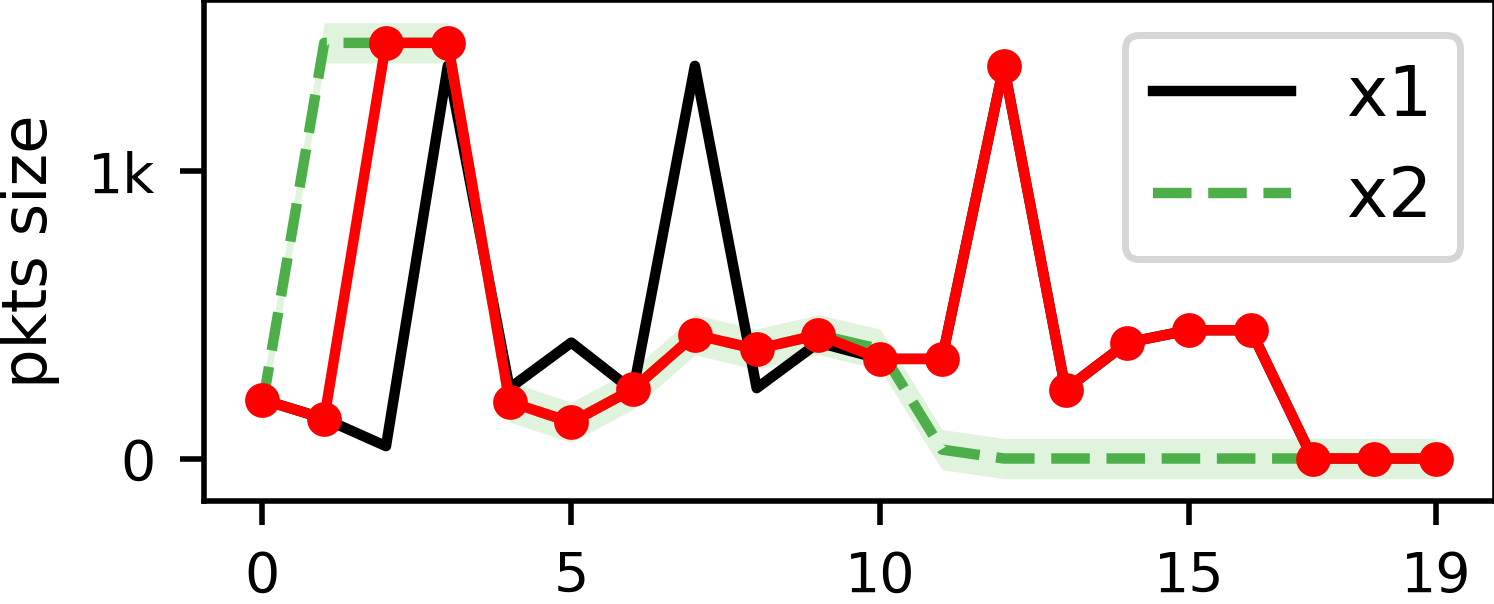}
\end{minipage}
\\
\hline
Packet Loss~\cite{horowicz2022miniflowpic} &
Remove values in a random time range (as if packets were not received)
\begin{detailbox}
\emph{Details:}
Defining $\Delta$ as time to observe the first $T$ packets,
sample $\delta\!\sim\!U[0, \Delta]$ and 
remove values across all features
in the interval $\delta\pm(10 \alpha + 5)$.
Then recompute the IAT and pad with zeroes
at the end (if needed).
\end{detailbox}
&       
\begin{minipage}{0.22\textwidth}
\centering $\delta\pm10$s $ \rightarrow t\in[10,14]$\\
\includegraphics[width=\linewidth]{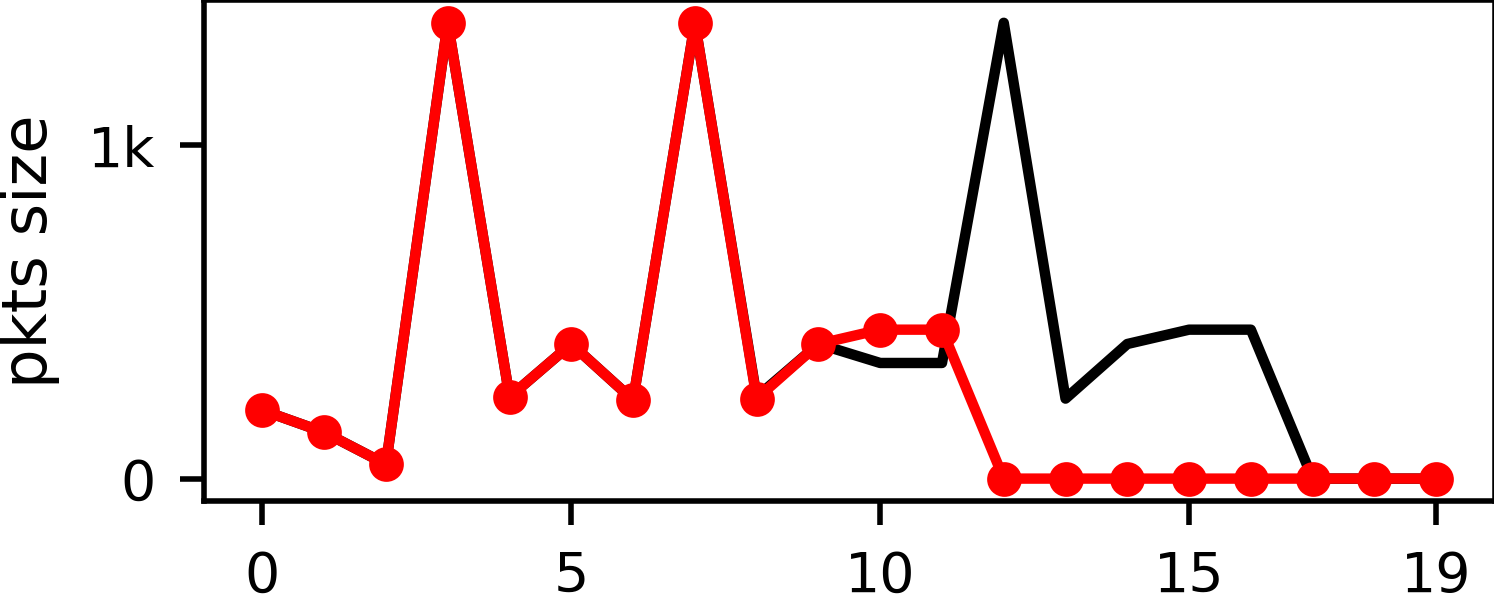}
\end{minipage}
\\
\hline
Translation  &   
Move a segment to left ($\approx$pkt drop) or the right ($\approx$ pkt dup/retran)
\begin{detailbox}
\emph{Details:}
Define $N\!\!=\!\!1\!\!+\!\!\textrm{arg max}_i \{a_i \le \alpha\}$ where \mbox{$a_i \in \{0.15, 0.3, 0.5, 0.8\}$} and sample $n\!\!\sim\!\!U[1, N]$. 
Then, sample a direction $b\!\!\in\!\!\{\emph{left}, \emph{right}\}$ and
a starting point \mbox{$t\!\!\sim\!\!U[0,T]$}:
If $b=left$, left shift each feature values $n$ times starting from $t$ and replace
shifted values with zero; if $b=right$, right shift each feature values $n$ times
starting from $t$ and replace shifted values with the single value $x_{(d,t)}$
\end{detailbox}
&     
\begin{minipage}{0.22\textwidth}
\centering $t=0, n = 1, b=right$\\
\includegraphics[width=\linewidth]{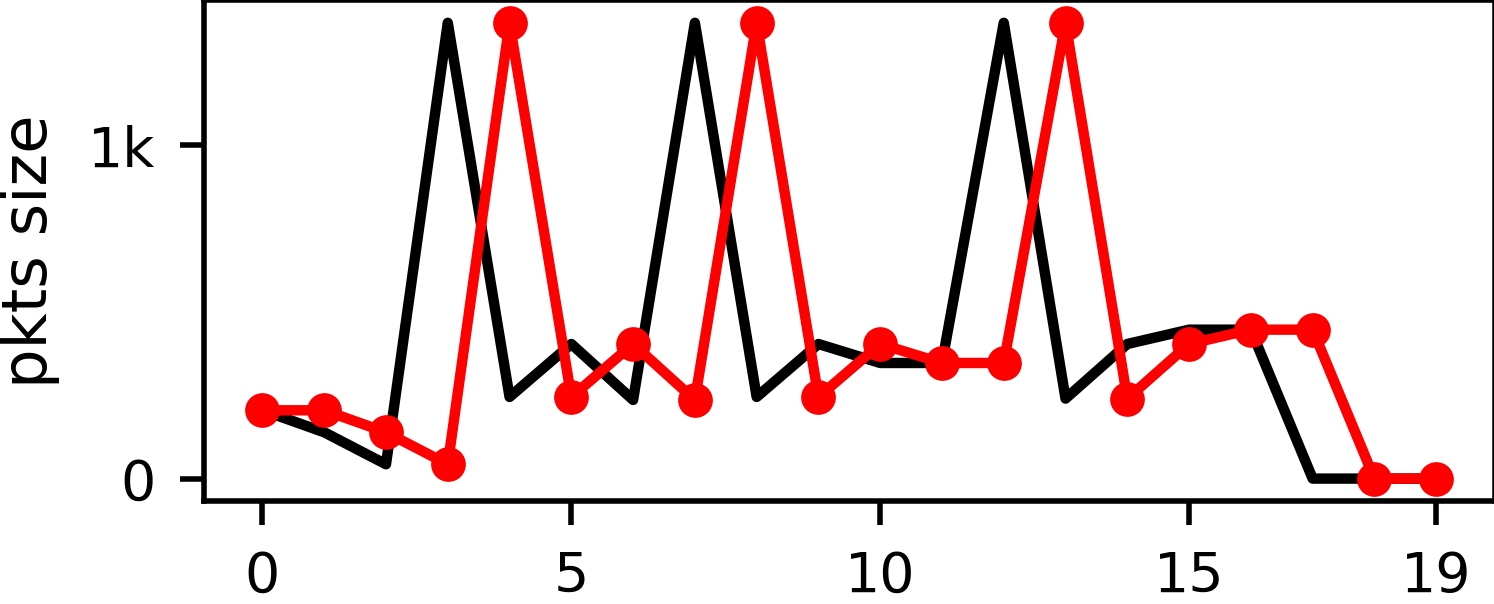}
\end{minipage}
\\
\hline
Wrap~\cite{Johannes2022CT_TS_Applied_Soft_Computing}  &   
Mixing interpolation, drop and no change
\begin{detailbox}
\emph{Details:}
Compose a new sample $\bf x'$ by manipulating each $x_{(:, t)}$ based on 
three options with probabilities
$P_{interpolate} = P_{discard} = 0.5\alpha$
and $P_{nochange}=1-\alpha$.
If ``nochange'' then keep $x_{(:, t)}$;
if ``interpolate'' then keep $x_{(:, t)}$ and
$x_{(:, t)}=0.5(x_{(:, t)} + x_{(:, t+1)})$;
if ``nochange'' then do nothing.
Stop when $|\bf x'|=T$ or apply tail padding (if needed).
\end{detailbox}
&       
\begin{minipage}{0.22\textwidth}
\includegraphics[width=\linewidth]{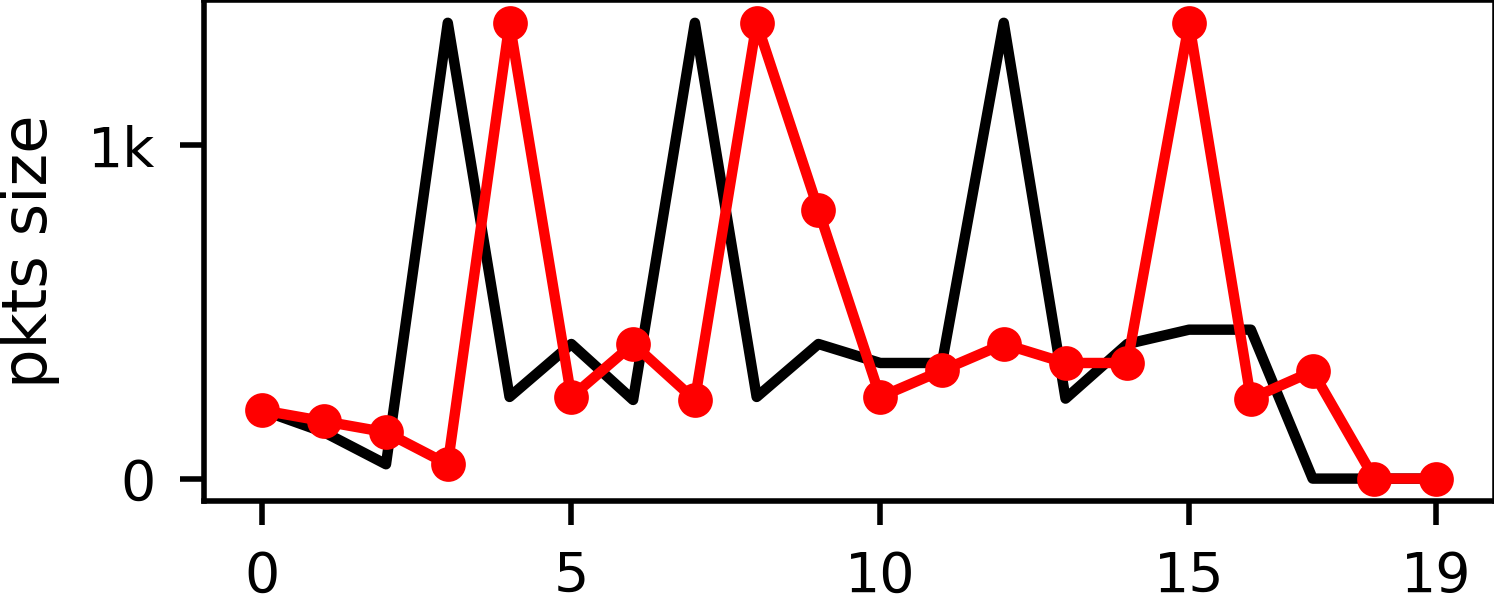}
\end{minipage}
\\
\hline
Permutation~\cite{Eldele2021ts_strong_weak_ct}
&   
Segment the time series and reorder the segments
\begin{detailbox}
\emph{Details:}
Define $N\!\!=\!\!2\!\!+\!\!\textrm{arg max}_i \{a_i \le \alpha\}$ where \mbox{$a_i\!\!\in\!\!\{0.15, 0.45, 0.75, 0.9\}$}, a sample $n\!\sim\!U[2, N]$ and split the range [0:T-1]
into $n$ segments of random length.
Compose a new sample $\bf x'$ by concatenating
$x_{(:, t)}$ from a random order of segments
\end{detailbox}
&       
\begin{minipage}{0.22\textwidth}
$n=3,$\\
\mbox{${\bf x'}\!=\![7\!\!:\!\!13]\!\cup\![0\!\!:\!\!6]\!\cup\![14\!\!:\!\!19]$}
\includegraphics[width=\linewidth]{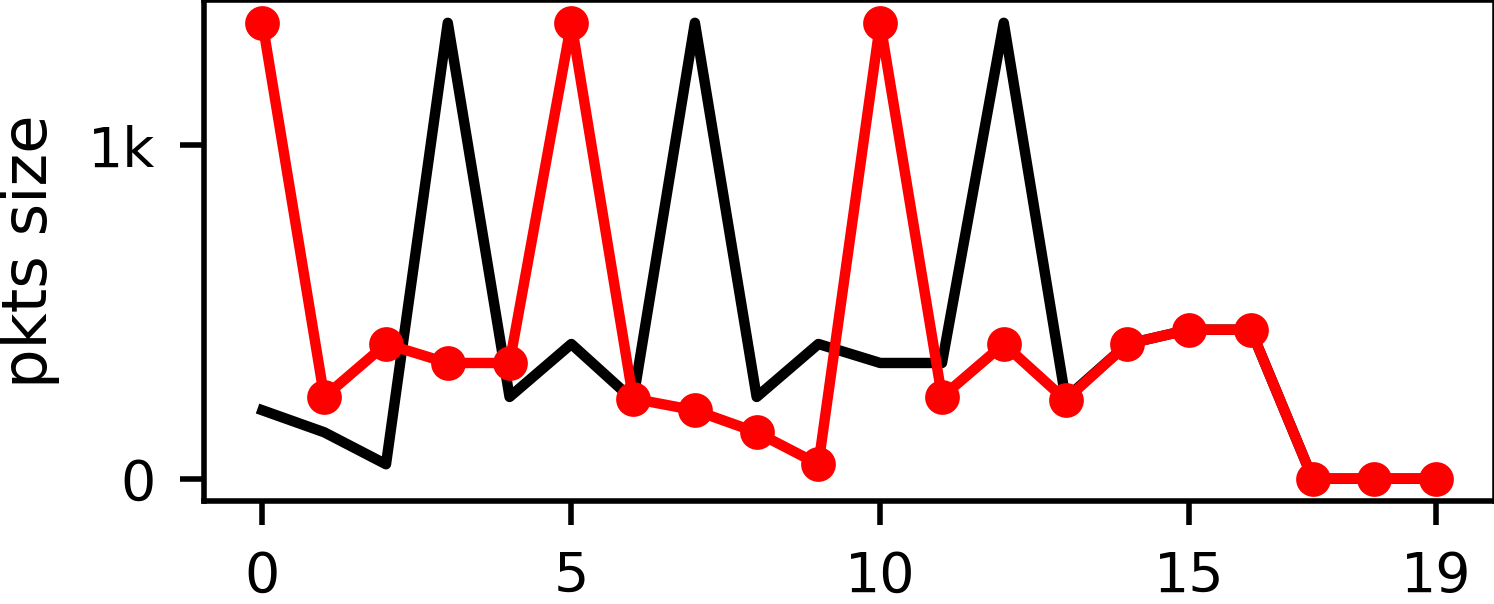}
\end{minipage}
\\
\hline
Dup-RTO~\cite{Xie2023DACT4TC} &
Mimic TCP pkt retrans due to timeout by duplicating values
\begin{detailbox}
\emph{Details:}
Duplicating a range of packets according
to a ${\rm Bernoulli}(p=0.1\alpha)$ 
(see Algo. 1 in~\cite{rosettareport})
\end{detailbox}
&       
\begin{minipage}{0.22\textwidth}
\includegraphics[width=\linewidth]{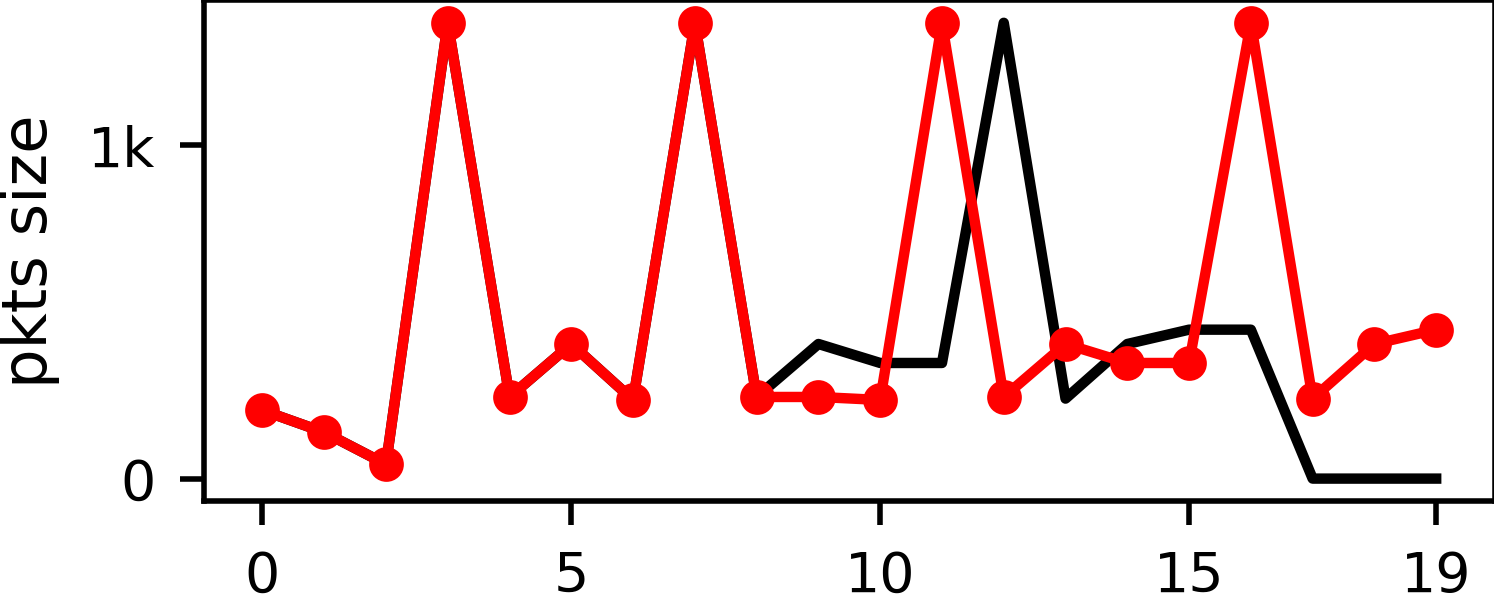}
\end{minipage}
\\
\hline
Dup-FastRetr~\cite{Xie2023DACT4TC}  & 
Mimic TCP fast retrans by duplicating values
\begin{detailbox}
\emph{Details:}
Duplicating one packet according
to a ${\rm Bernoulli}(p=0.1\alpha)$ 
(see Algo. 2 in~\cite{rosettareport})
\end{detailbox}
&       
\begin{minipage}{0.22\textwidth}
\includegraphics[width=\linewidth]{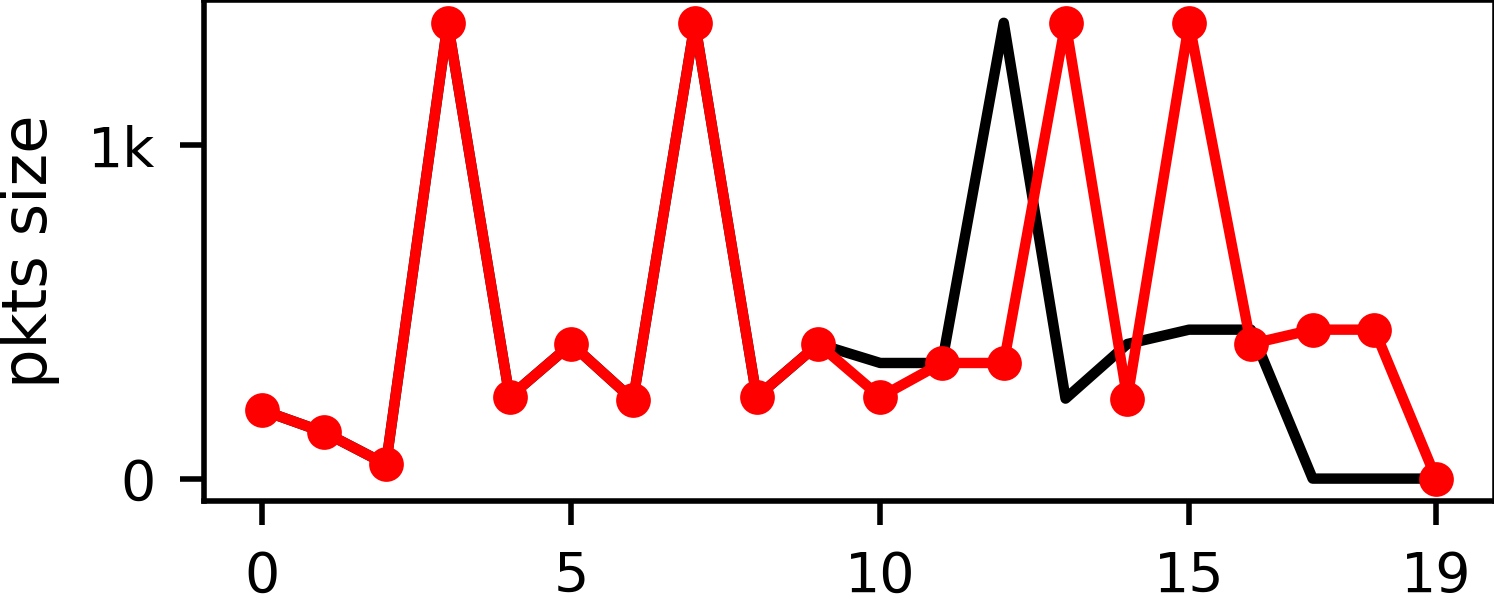}
\end{minipage}
\\
\hline
Perm-RTO~\cite{Xie2023DACT4TC}  &
Mimic TCP pkt retrans due to timeout by permuting values
\begin{detailbox}
\emph{Details:}
Delaying a range of packets according
to a ${\rm Bernoulli}(p=0.1\alpha)$ 
(see Algo. 3 in~\cite{rosettareport})
\end{detailbox}
&       
\begin{minipage}{0.22\textwidth}
\includegraphics[width=\linewidth]{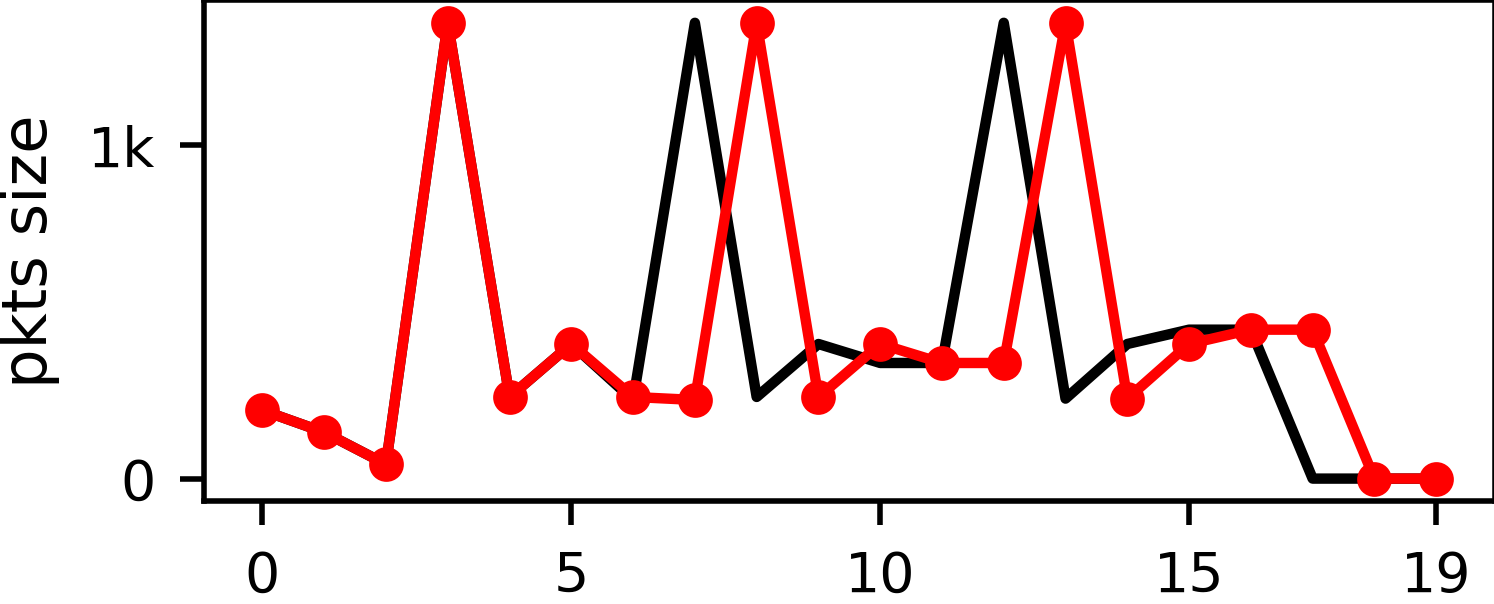}
\end{minipage}
\\
\hline
Perm-FastRetr~\cite{Xie2023DACT4TC}  &
Mimic TCP fast retrans by permuting values
\begin{detailbox}
\emph{Details:}
Delaying one packet according
to a \mbox{${\rm Bernoulli}(p=0.1\alpha)$} 
(see Algo. 4 in~\cite{rosettareport})
\end{detailbox}
&       
\begin{minipage}{0.22\textwidth}
\includegraphics[width=\linewidth]{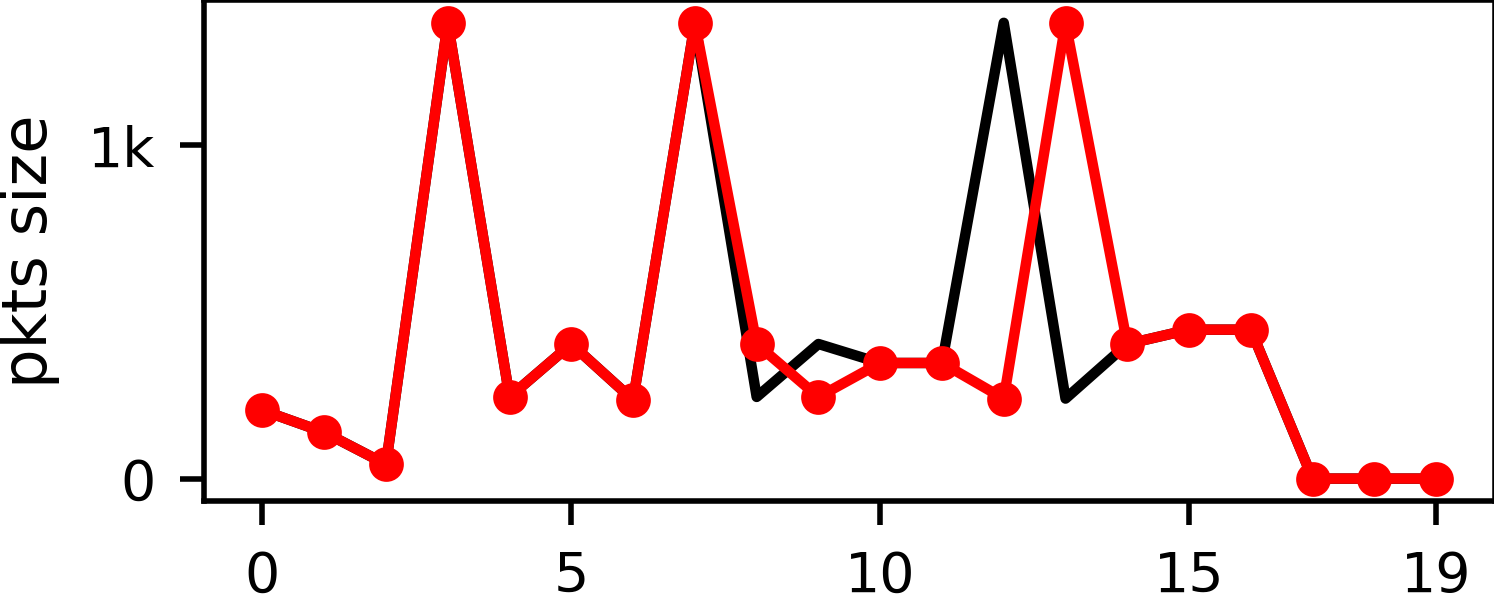}
\end{minipage}
\\
\bottomrule
\end{longtable}
\raggedright
\vspace{-1em}
In the figures, black solid lines \rule[0.1em]{1em}{2pt} for original samples, \textcolor{red}{red $\CIRCLE$ lines} for augmented samples.
\end{center}
\endgroup

\clearpage
\subsection{Benchmarking hand-crafted DA (G1)}\label{sec: MethodsAugmentations}

Figure~\ref{fig:input-sample} sketches a typical TC input $\rm {\bf x}$,
i.e., a multivariate time series with $D$ dimensions (one for each packet feature) each having
$T$ values (one for each packet) while $x_{(d,t)}$ is the value of $\rm {\bf x}$ 
at coordinates $(d, t)$ where $d\in\{0..D-1\}$ and $t\in\{0..T-1\}$.
In particular, in this work, we consider $D=3$ packet features,
namely packet size, direction, and Inter Arrival Time (IAT), and the first
$T=20$ packets of a flow.
We also define $\rm {\bf x'} = {\rm Aug}(\rm {\bf x}, \alpha)$ as an augmentation, i.e., 
the transformation ${\rm\bf x'}$ of sample $\rm {\bf x}$ is subject to a \emph{magnitude} $\alpha \in ]0,1[$ 
controlling the intensity of the transformation (1 = maximum modification).

\paragraph{\bf Augmentations pool.}
In this study, we considered a set $\mathcal{A}$ of 18 augmentation functions. These 
functions can be categorized into 3 families: 
5 \emph{amplitude} transformations, which introduce different
type of jittering to the feature values (Table~\ref{tab:augs_amplitude});
2 \emph{masking} transformations, which force certain feature values to zero 
(Table~\ref{tab:augs_mask}); and 11 \emph{sequence} transformations, which modify the order of 
feature values (Table~\ref{tab:augs_order}).
It is important to note that, given a sample $\rm {\bf x}$, amplitude augmentations are solely 
applied to either packet size or IAT while packet direction is never altered since the latter 
is a binary feature and does not have amplitude (i.e., it can be $-1$ or $1$).
On the contrary, masking and sequence augmentations are applied to all features in parallel (e.g., if a 
transformation requires to swap $t=1$ with $t=6$, all features are swapped accordingly $x_{(i, 1)} \leftrightarrow x_{(i, 6)}$ for $\forall i\in\{0..D-1\}$). 
For each augmentation, Tables~\ref{tab:augs_amplitude}-\ref{tab:augs_order} report a reference example annotating its parametrization (if any).

By adopting such a large pool of augmentations our empirical campaign offers several 
advantages. First, we are able to investigate a broader range of design possibilities 
compared to previous studies. Second, it enables us to contrast different 
families and assess if any of them is more prone to disrupt class semantics. Considering the 
latter, TC literature~\cite{horowicz2022miniflowpic,rezaei2019ICDM-ucdavis,Xie2023DACT4TC} 
predominantly investigate sequence transformations (typically acting only on packet timestamp)
with only~\cite{Xie2023DACT4TC} experimenting with masking and amplitude variation, yet
targeting scenarios where models are exposed to data shifts due to maximum segment size (MSS) 
changes, i.e., the network properties related to the training set are different from
the ones of the test set.

\paragraph{\bf Augmentations magnitude.}
As described in Tables~\ref{tab:augs_amplitude}-\ref{tab:augs_order}, 
each augmentation has some predefined static parameters\footnote{These 
parameters are tuned via preliminary investigations.} while 
the magnitude $\alpha$ is the single hyper-parameter 
controlling random sampling mechanisms contributing to defining the final
transformed samples. To quantify augmentations sensitivity to $\alpha$, we contrast two
scenarios following CV literature practice: a static value of $\alpha=0.5$ and a uniformly sampled value $\alpha \sim U[0,1]$ extracted for each augmented sample.

\paragraph{\bf Datasets size and task complexity.}
Supervised tasks, especially when modeled via DL, benefit
from large datasets. For instance, as previously mentioned, some CV literature
pretrains generative models on large datasets and use those models
to obtain auxiliary training data for classification tasks.
While data availability clearly plays a role, at the same time the task
complexity is equivalently important---a task with just a few classes
but a lot of data does not necessarily yield higher accuracy than
a task with more classes and less data.
To understand how augmentations interplay with these dynamics, it is 
relevant to evaluate augmentations across 
datasets of different sizes and number of classes.

\begin{figure}[t]
\centering
\includegraphics[width=\textwidth]{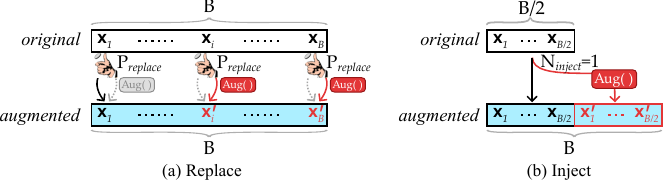}
\caption{Training batch creation policies.\label{fig:batch_creation}
}
\end{figure}

\subsection{Training batches composition (G2)
\label{sec:method_batching}
}

In order to mitigate any undesirable shifts introduced by artificial samples, it is necessary 
to balance original and augmented samples. Yet, the way original and augmented samples are combined to 
form the augmented training set is a design choice. For instance, 
in TC literature, \cite{horowicz2022miniflowpic,rezaei2019ICDM-ucdavis} 
augment the data before starting the training, while
\cite{Xie2023DACT4TC} augments mini-batches during the training process.
In this work, we apply augmentation samples to a training mini-batch of size $B$, 
with the two policies sketched in Fig.~\ref{fig:batch_creation}. \emph{Replace}
substitutes an original sample $\rm {\bf x_i}$ with its augmentation
by sampling from a Bernoulli(P=$P_{replace}$) random variable---during one training epoch,
approximately a $P_{replace}$ fraction of the original data is ``hidden''. 
Instead, \emph{Inject} increases the batch size by
augmenting each sample $N_{inject}$ times (e.g., in Fig.~\ref{fig:batch_creation}
the original batch size is doubled by setting $N_{inject}=1$).

\subsection{Latent space geometry (G3)}
Augmented samples play a crucial role in model training, just like the original training samples from 
which they are derived. To understand the impact of augmentations on the improvement or detriment of 
classification performance, we propose to examine the latent space of the classifier. In order to conduct a comprehensive 
analysis, we need to consider two aspects applicable to any supervised classification 
task.

\vspace{3pt}
\noindent \textbf{Augmentation-vs-Test.}
ML methods operate on the assumption that 
training data serves as a ``proxy'' for test samples, i.e., the patterns
learned on training data ``generalize'' to testing data
as the two sets of data resemble each other properties.
In this context, augmentations can be considered as a means 
for fostering data generalization by incorporating samples that resemble even more testing 
data compared to what is available in training data. However, it is important to empirically quantify this effect 
by measuring, for instance,
the distance between augmented and test samples.
In other words, we aim to quantify up to which extent
augmented samples are better at mimicking test samples
compared to the original data.

\vspace{3pt}
\noindent \textbf{Augmentation-vs-Train.}
The performance of a feature extractor greatly depends on how well the feature extractor 
separates different classes in the latent space. Data augmentations play a role in  
shaping intra/inter-class relationships created by the feature extractor in the latent space.
For instance, an augmentation that generates samples far away from
the region of a class can \emph{disrupt class semantics}---the
augmentation is introducing a new behavior/mode making it hard for the classifier to be effective.
At the same time, however, expanding the region of a class can be a beneficial 
design choice---augmentations that enable a better definition of class
boundaries simplify the task of the classifier. 
Understanding such dynamics requires empirical observations, 
for instance, by comparing the distance between original training data
and augmented data. In other words, we aim to verify
if augmentations yielding good performance are in a
``sweet spot'': they create samples that are neither 
too close (i.e., introduce too little variety) nor too far (i.e., disrupt class semantics) from original samples.

\subsection{Combining augmentations (G4)\label{sec:combining_aug}}
To address \textbf{G1}, each trained model is associated
to an individual augmentation.
However, in CV it is very common to combine multiple augmentations~\cite{chen2020simclr}.
Hence, we aim to complement \textbf{G1} by measuring the performance
of three different policies augmenting mini-batches based on a 
set $\mathcal{A'} \subset \mathcal{A}$ composed of top-performing augmentations based on the \textbf{G1} benchmark: the 
\emph{Ensemble} policy uniformly samples one of the augmentations in 
$\mathcal{A'}$ independently for each mini-batch sample; the 
\emph{RandomStack} policy randomly shuffles $\mathcal{A'}$ 
independently for each mini-batch sample before applying all augmentations; finally, the
\emph{MaskedStack} policy uses a predefined order for $\mathcal{A'}$
but each augmentation is associated to a masking probability, i.e.,
each sample in the mini-batch independently selects a subset of augmentations
of the predefined order.

\section{Experimental settings
\label{sec:experimental_settings}} 

\subsection{Datasets}

To address our research goals we considered the datasets summarized in Table~\ref{tab:datasets}.

\vspace{3pt}
\noindent\MIRAGEA~\cite{aceto2019mirage}
is a \emph{public} dataset
gathering traffic logs from $20$ popular Android apps\footnote{Despite being advertised with having traffic from 40 apps, the \emph{public} version of the dataset only contains 20 apps.}
collected at  
the ARCLAB laboratory of the University of Napoli Federico II.
Multiple measurement campaigns were operated 
by instrumenting $3$ Android devices 
handed off to $\approx$$300$ volunteers (students and researchers)
for interacting with the selected apps for short sessions.
Each session resulted in a pcap file and an \texttt{strace} log 
mapping each socket to the corresponding Android application name.
Pcaps were then post-processed to obtain bidirectional flow logs by grouping all packets belonging to the same 5-tuple (srcIP, srcPort, dstIP, dstPort, L4proto) and extracting both aggregate metrics (e.g., total bytes, packets, etc.), per-packet time series (packet size, direction, TCP flags, etc.), raw packets payload bytes
(encoded as a list of integer values) and mapping a ground-truth label by means of the \texttt{strace} logs.

\arrayrulecolor{black}
\begin{table}[t]
    \centering
    \footnotesize
    \caption{Summary of datasets properties.}
    \label{tab:datasets}
\begin{tabular}{
    l 
    @{$\,\,\,\,\,\,\,$}c 
    @{$\,\,\,\,\,\,\,$}r
    @{$\,\,\,\,$}r
    @{$\,\,\,\,$}r 
    @{$\,\,\,\,$}r 
    @{$\,\,\,\,$}c
    @{$\,\,\,\,$}r
    }
\toprule
\multirow{2}{*}{\bf Name}
& \multirow{2}{*}{\bf Classes}
& \multirow{2}{*}{\bf Curation}
& \multicolumn{4}{c}{\bf Flows per-class}
& \bf Pkts
\\
&&
& \emph{all}
& \emph{min} 
& \emph{max}
& \emph{$\rho$}
& \emph{mean}
\\
\midrule
\multirow{2}{*}{\MIRAGEA~\cite{aceto2019mirage}}
    &  \multirow{2}{*}{20} 
    & \multicolumn{1}{c}{none}
      &                            122~k &      1,986 &     11,737 &        5.9 &         23\\
    &   
    & $>$\emph{10pkts}&             64~k &      1,013 &      7,505 &        7.4 &         17
\\
\midrule
\multirow{2}{*}{\MIRAGEB\cite{guarino2021mirage}}
     & \multirow{2}{*}{9}
     & \multicolumn{1}{c}{none}
       &                              59~k &      2,252 &     18,882 &        8.4 &      3,068\\
  & & $>$\emph{10pkts}&               26~k &        970 &      4,437 &        4.6 &      6,598
\\
 \midrule
    \HOMECAMPUS
     & 100
     & \multicolumn{1}{c}{none}        &     2.9~M &        501,221 &      5,715 &       87.7 &        2,312
\\
\bottomrule
\multicolumn{8}{p{29em}}{
\scriptsize
$\rho$ : ratio between max and min number of flows per-class---the larger the value, the higher the imbalance;
}\\
\end{tabular}
\\
\scriptsize
\end{table}

\vspace{3pt}
\noindent\MIRAGEB~\cite{guarino2021mirage} 
is another \emph{public} dataset
collected by the same research team and with the same
instrumentation as \MIRAGEA which targets 9
video meeting applications used to perform
webinars (i.e., meetings with multiple attendees
and a single broadcaster), audio calls (i.e., meetings with
two participants using audio-only), video calls (i.e., meetings with two participants 
using both audio and video), and video conferences, (i.e., meetings
involving more than two participants broadcasting audio and video).

\vspace{3pt}
\noindent\HOMECAMPUS is instead a \emph{private}\footnote{Due to NDA we
are not allowed to share the dataset.} 
dataset collected by monitoring network flows
from vantage points deployed in residential
access and enterprise campus networks.
For each flow, the logs report multiple aggregate metrics
(number of bytes, packets, TCP flags counters,
round trip time statistics, etc.), and 
the packet time series of packet size, direction
and IAT for the first 50 packets of each flow.
Moreover, each flow record is also enriched with an 
application label provided by a commercial DPI software
directly integrated into the monitoring solution
and supporting hundreds of applications and services.

\vspace{5pt}
\noindent{\bf Data curation.}
Table~\ref{tab:datasets} compares different dataset properties. 
For instance, \MIRAGEA and \MIRAGEB are quite different from each other
despite being obtained via the same platform.
Specifically, \MIRAGEA gathers around 2$\times$ more flows than \MIRAGEB but
those are 100$\times$ shorter. 
As expected, all datasets are subject to class imbalance 
measured by $\rho$, i.e., the ratio between maximum and minimum
number of samples per class. However, \HOMECAMPUS exhibits a larger class imbalance
with respect to the other two datasets. Last, while \HOMECAMPUS did not require 
specific pre-processing, both \MIRAGEA and \MIRAGEB required
a curation to remove \emph{background traffic}---flows created by netd deamon, SSDP, Android 
google management services and other services unrelated to the target Android apps---and 
flows having less than 10 packets.

\vspace{5pt}
\noindent{\bf Data folds and normalization.}
As described in Sec.~\ref{sec: MethodsAugmentations}, 
each flow is modeled via a multivariate time series 
$\rm {\bf x}$ consisting of $D=3$ features (packets size, 
direction, and IAT) related to the first $T=20$
packets (applying zero padding in the tail where needed). 
From the curated datasets we created 80 random 70/15/15
train/validation/test folds.
We then processed each train+val split to extract
statistics that we used for normalizing the data and to drive the augmentation process.
Specifically, we computed both per-coordinate $(d, t)$ and global (i.e., flattening
all flows time series into a single array) mean and standard deviation
for each class---these statistics provided us the $\sigma_{(d,t)}^y$ and $\sigma_{(:,t)}^y$ 
needed for the augmentations (see Fig.~\ref{fig:input-sample} and 
Tables~\ref{tab:augs_amplitude}-\ref{tab:augs_order}).
For IAT, we also computed the global 99th percentile across all classes $q_{iat}^{99}$.
Given a multi-variate input $\rm {\bf x}$, 
we first clip packet size values in the range $[0,1460]$ and IAT values in the range [1e-7, $q_{iat}^{99}$]. 
Due to high skew of IAT distributions, we also log10-scaled the IAT
feature values.\footnote{We did not log-scale packet sizes values as we found this can reduce accuracy
based on preliminary empirical assessments.}
Last, all features are standardized to provide values $x_{(d,t)}\in[0,1]$.

\vspace{5pt}
\noindent\textbf{Model architecture and training.} 
We rely on a 1d-CNN based neural network architecture with
a backbone including 2 ResNet blocks followed by a linear head
resulting in a compact architecture of $\approx$100k parameters.
(see Fig.~\ref{fig:model} and Listing~\ref{lst:net} in the appendix for details).
Models are trained for a maximum of 500 epochs with a batch size B=1,024 
via an AdamW optimizer with a weight decay of 0.0001 and a 
cosine annealing learning rate scheduler initialized at 0.001.
Training is subject to early stopping by monitoring if the validation accuracy
does not improve by 0.02 within 20 epochs.
We coded our modeling framework using PyTorch and PyTorch Lightning and ran our 
modeling campaigns on Linux servers equipped with multiple NVIDIA Tesla V100 GPUs.
We measured the classification performance via the weighted F1 score
considering a reference baseline where training is not subject to augmentations.

\section{Results
\label{sec:results}
}

In this section, we discuss the results of our modeling
campaigns closely following the research goals introduced
in Sec.~\ref{sec:goals}.

\subsection{Augmentations benchmark (G1)
\label{sec:results_rank}
}

We start by presenting the overall performance
of the selected augmentations. Specifically,
Table~\ref{tab:g1_benchmark} collects results
obtained by applying augmentations via Inject
with $N_{inject}=1$ (i.e., 
each original sample is augmented once)\footnote{Since we 
train the reference baseline with a batch size B=1024,
when adding augmentations we instead adopt B=512 (which
doubles via injection). 
}
and sampling uniformly the magnitude $\alpha \sim U[0,1]$.
Table~\ref{tab:g1_benchmark} shows
the average weighted F1 score across 80 
runs and related 95th-percentile confidence intervals.

\begin{table}[t]
\centering
\caption{Augmentations benchmark \textbf{(G1).}
\label{tab:g1_benchmark}
}
\label{tab:aug_MIRAGEA}
\footnotesize
\begin{tabular}{
    l
    l
    r
    @{$\:\:\:\:\:\:\:\:$}
    r
    @{$\:\:\:\:\:\:\:\:$}
    r
}
\toprule
& \bf Augmentation & \MIRAGEA & \MIRAGEB & \HOMECAMPUS\\
\cmidrule(r){1-2}
\cmidrule(r){3-5}
\bf Baseline
& None                 &  75.43\tinytiny{.10} &  94.92\tinytiny{.07} & 92.43\tinytiny{.33} \\
\cmidrule(r){1-2}
\cmidrule(r){3-5}
\multirow{4}{*}{\bf Amplitude}
& Constant WrapUp      &       \WORSTPERF      0.61\tinytiny{.12}  &             0.36\tinytiny{.09} &-0.02\tinytiny{.15}\\
& Gaussian Noise       &             0.89\tinytiny{.11}  &  \WORSTPERF 0.24\tinytiny{.09} &0.15\tinytiny{.14}\\
& Gaussian WrapUp      &             1.01\tinytiny{.13}  &             0.74\tinytiny{.09} &0.24\tinytiny{.12}\\
& Spike Noise          &             1.66\tinytiny{.12}  &             0.91\tinytiny{.09} &0.93\tinytiny{.13}\\
& Sine WrapUp          &             0.63\tinytiny{.11}  &             \WORSTPERF0.25\tinytiny{.09} &-0.06\tinytiny{.16}\\
\cmidrule(r){1-2}
\cmidrule(r){3-5}
\multirow{2}{*}{\bf Masking} 
& Bernoulli Mask       &             2.55\tinytiny{.12}  &             1.29\tinytiny{.09} &\BESTPERF 1.25\tinytiny{.16}\\
& Window Mask          &             2.37\tinytiny{.13}  &             1.08\tinytiny{.09} &\BESTPERF 1.18\tinytiny{.16}\\
\cmidrule(r){1-2}
\cmidrule(r){3-5}
\multirow{11}{*}{\bf Sequence}
& CutMix               &             2.65\tinytiny{.13}  &             1.40\tinytiny{.10} &\WORSTPERF-0.21\tinytiny{.10}\\
& Dup-FastRetr         &             3.23\tinytiny{.13}  &             1.56\tinytiny{.09} &0.83\tinytiny{.15}\\
& Dup-RTO              &             2.89\tinytiny{.13}  &             1.33\tinytiny{.09} &0.91\tinytiny{.15}\\
& Horizontal Flip      &  \WORSTPERF-0.71\tinytiny{.11}  &  \WORSTPERF-0.52\tinytiny{.09} &	\WORSTPERF -0.88\tinytiny{.15}\\
& Interpolation        &  \WORSTPERF 0.44\tinytiny{.12}  &             0.53\tinytiny{.10} &\WORSTPERF-0.61\tinytiny{.14}\\
& Packet Loss          &             0.88\tinytiny{.12}  &             0.66\tinytiny{.09} &0.60\tinytiny{.22}\\
& Permutation          &          \BESTPERF    3.67\tinytiny{.13}  &           \BESTPERF   1.97\tinytiny{.09} &0.89\tinytiny{.08}\\
& Perm-RTO             &             3.15\tinytiny{.12}  &             1.54\tinytiny{.09} &0.88\tinytiny{.12}\\
& Perm-FastRetr        &             2.11\tinytiny{.12}  &             1.00\tinytiny{.09} &0.74\tinytiny{.26}\\
& Translation          &  \BESTPERF  4.40\tinytiny{.13}  &   \BESTPERF 2.02\tinytiny{.09} &\BESTPERF 0.95\tinytiny{.15}\\
& Wrap                 &  \BESTPERF  4.11\tinytiny{.13}  &   \BESTPERF 2.09\tinytiny{.08} &0.57\tinytiny{.12}\\
\bottomrule
\multicolumn{5}{l}{
\scriptsize The top-3 best and worst augmentations are color-coded.
}
\end{tabular}
\end{table}

\begin{figure}[t]
  \centering
  \hspace{-3mm}
 \begin{subfigure}{0.51\textwidth}
    \includegraphics[width=\textwidth]{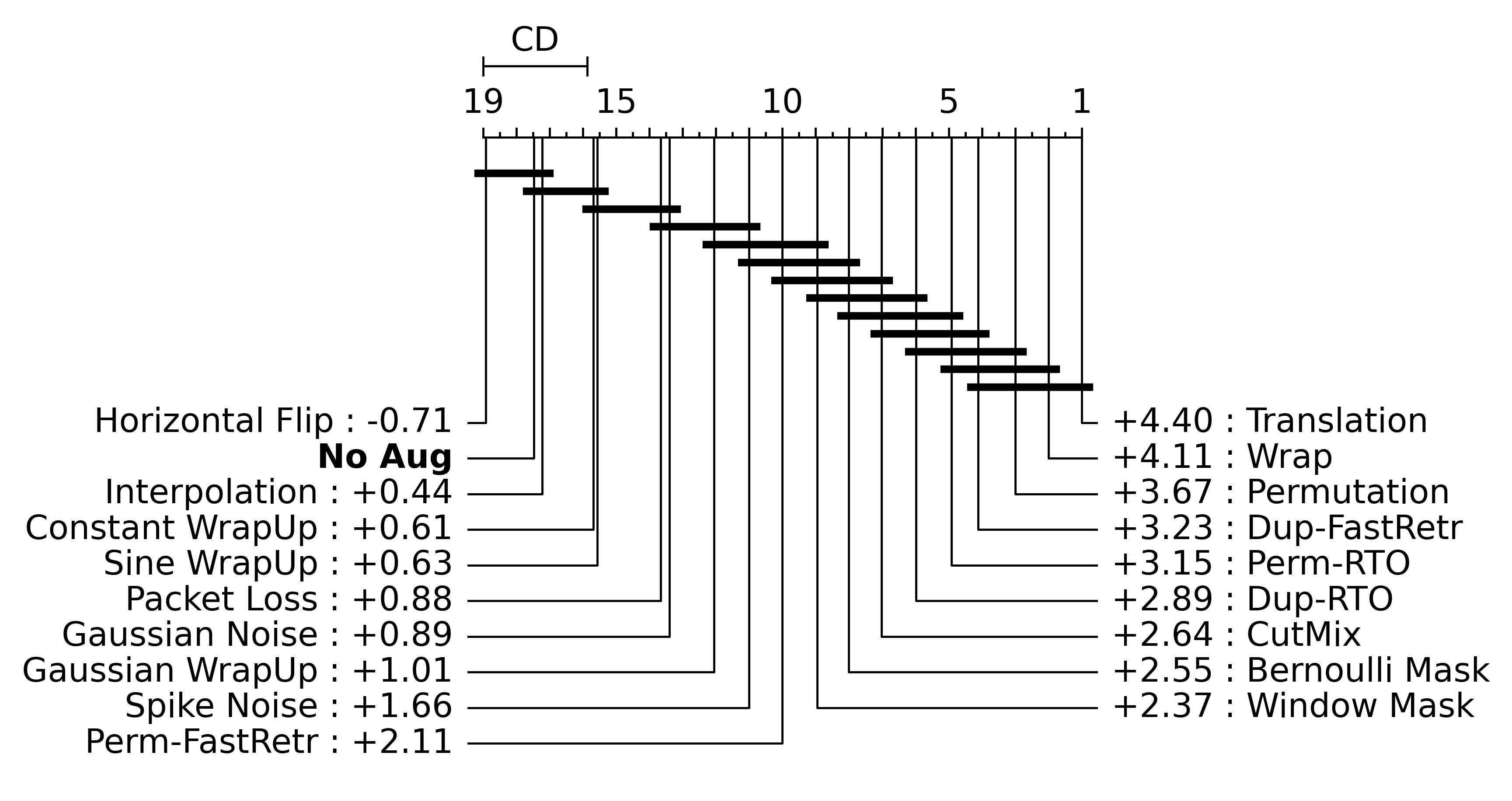} 
    \caption{\MIRAGEA}%
  \end{subfigure}
\hspace{-3mm}
   \begin{subfigure}{0.51\textwidth}
    \includegraphics[width=\textwidth]{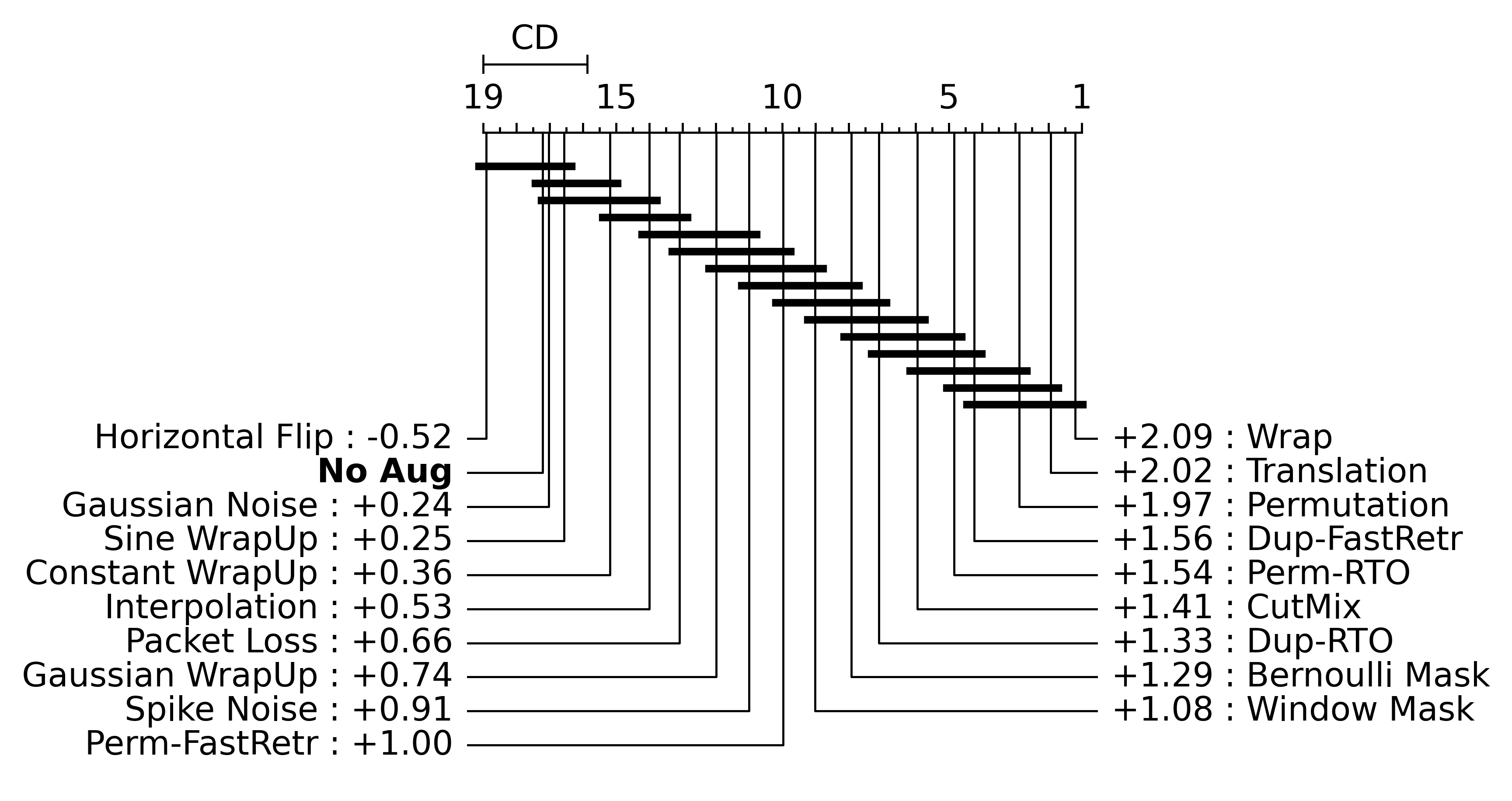} 
    \caption{\MIRAGEB}%
  \end{subfigure}

\caption{Augmentations rank and critical distance \textbf{(G1).}
\label{fig:g1_critical_distance}
} 
\end{figure}

\vspace{5pt}
\noindent {\bf Reference baseline.}
We highlight that our reference baseline
performance for \MIRAGEA and \MIRAGEB are \emph{qualitatively}
aligned with previous literature that used those datasets.
For instance, Table 1 in \cite{guarino2021mirage} reports
a weighted F1 of 97.89 for a 1d-CNN model
when using the first 2,048 payload bytes as input for \MIRAGEB; 
Figure 1 in \cite{bovenzi2021tma} instead shows
a weighted F1 of $\approx$75\% for 100 packets time series
input for \MIRAGEA.
Notice however that since these studies use training configurations
not exactly identical to ours, 
a direct comparison with our results should be taken with caution. 
Yet, despite these differences, we confirm 
\MIRAGEA to be a more challenging classification task compared to \MIRAGEB.
However, we argue that such a difference is unlikely depending only
on the different number of classes (\MIRAGEA has 20 classes
while \MIRAGEB only 9). This is evident by observing that \HOMECAMPUS 
yields very high performance despite having 10$\times$ more 
classes than the other two datasets. We conjecture instead 
the presence of ``cross-app traffic'' such as flows 
generated by libraries/services common across multiple apps
from the same provider (e.g., apps or services provided by Google or Facebook)
and/or the presence of ads traffic,\footnote{
\MIRAGEB focuses on video meeting apps which
are all from different providers and ads free by design.
}
but the datasets raw data is not sufficiently detailed to investigate
our hypothesis.

\noindent \textit{\textbf{Takeaways.}
While the classification tasks complexity is well captured by models
performance, it does not necessarily relate to the number of classes 
or dataset size. These effects are visible 
only when studying multiple datasets at once,
but unfortunately a lot of TC studies focus on individual datasets.
}

\vspace{5pt}
\noindent {\bf Augmentations rank.}
Overall, all augmentations are beneficial except for Horizontal Flip
which, as we shall see in Sec.~\ref{sec:results_latent_space}, breaks class semantics. As expected, not all augmentations provide the same gain and their effectiveness may vary across datasets. Specifically, \emph{sequence} and \emph{masking} better suit our TC tasks.

For a more fine-grained performance comparison, we complement
Table~\ref{tab:g1_benchmark} results by analyzing augmentations
rank via a critical distance by following the procedure described in
\cite{demvsar2006statistical}. Specifically,
for each of the 80 modeling runs we first
ranked the augmentations from best to worst (e.g., if augmentations A, B, and C yield a weighted F1 of 0.9, 0.7, and 0.8, their associated rankings would be 1, 3, and 2) splitting ties using the average ranking of the group (e.g., if augmentations A, B, and C yield a weighted F1 of 0.9, 0.9 and 0.8, their associated rankings would be 1.5, 1.5, and 3). This process is then repeated across the 80 runs
and a global rank is obtained by computing the mean
rank for each augmentation. 
Last, these averages are compared pairwise using a post-hoc Nemenyi test
to identify which groups of augmentations are statistically equivalent. 
This decision is made using a Critical Distance 
$CD = q_{\alpha}\sqrt{k(k+1)/6N}$, 
where $q_{\alpha}$ is based on the Studentized range statistic divided by $\sqrt{2}$, $k$ is equal to the number of augmentations compared and $N$ is equal to the number of samples used. Results are then collected in Fig.~\ref{fig:g1_critical_distance} where
each augmentation is highlighted with its average rank (the lower the better)
and horizontal bars connect augmentations that are statistically equivalent.
For instance, while Table~\ref{tab:g1_benchmark} shows that
Translate is the best on average, Fig.~\ref{fig:g1_critical_distance}
shows that \{Translate, Wrap, Permutation, Dup-FastRetr\} are 
statistically equivalent. We remark that 
Fig.~\ref{fig:g1_critical_distance} refers to 
\MIRAGEA and \MIRAGEB but similar considerations hold for \HOMECAMPUS as well.

Recall that our training process is subject to an early stop mechanism.
Interestingly, we observed that augmentations yielding better performance
also present a longer number of training epochs 
(see Fig.~\ref{fig:g1_gain_vs_epochs} in Appendix).
This hints that effective augmentations foster 
better data representations extraction,
although some CV studies also show that 
early stopping might not necessarily be the best option
to achieve high accuracy in some scenarios. An 
in-depth investigation of these training mechanisms
is however out of scope for this paper.

\noindent\textit{\textbf{Takeaways.} 
Augmentations bring benefits that, 
in absolute scale, are comparable to what is observed in CV literature~\cite{muller2021trivialaugment}.
Our benchmark shows that TC sequencing and masking augmentations are better options
than amplitude augmentations.
This confirms previous literature that \emph{implicitly} discarded amplitude augmentations. 
Finally, despite performance ranks can suggest more performant augmentations
(e.g., Translation or Bernoulli mask), agreement between datasets seems more qualitative than punctual (e.g., masking is preferred to sequencing for \HOMECAMPUS, but the reverse is true for the other two datasets).
}

\vspace{5pt}
\noindent \textbf{Sensitivity to magnitude.}
Most of the augmentations we analyzed are subject
to a magnitude $\alpha$ hyper-parameter (see Tables~\ref{tab:augs_amplitude}-\ref{tab:augs_order}) that is randomly selected 
for the results in Table~\ref{tab:g1_benchmark}. 
To investigate the relationship between classification
performance and augmentation magnitude
we selected 3 augmentations among the top performing ones 
\{Translation, Wrap, Permutation\}
and three among the worst performing \{Gaussian Noise, Sine WrapUp, Constant WrapUp\}.\footnote{We excluded HorizontalFlip as it hurt performance and Interpolation
since it does not depend from a magnitude.}
For each augmentation, we performed 10 modeling runs 
using magnitude $\alpha = 0.5$ and we contrasted
these results with the related runs from the previous modeling campaign.
Specifically, by grouping all results we obtained a binary random-vs-static performance comparison which we
investigated through a Wilcoxon signed rank sum test
that indicated \emph{no statistical difference}, i.e., 
the selection of magnitude is not a distinctive factor
to drive the augmentation performance. 
The same conclusion holds true when
repeating the analysis 
for each individual augmentation
rather than grouping them together.

\noindent\textit{\textbf{Takeaways.} 
Although we do not observe any dependency on the 
augmentation magnitude $\alpha$, 
augmentations performance can still be affected by their tuning 
(as will be discussed further in Sec.~\ref{sec:results_latent_space}).
Unfortunately, this tuning process 
often relies on a trial-and-error process, making it challenging to operate manually.}

\subsection{Training batches composition (G2)
\label{sec:results_training_batches}
}

Correctly mixing original with augmented data is an important design choice. 

\begin{figure}[t]
  \centering
  
 \begin{subfigure}{0.495\textwidth}
    \includegraphics[width=\textwidth]{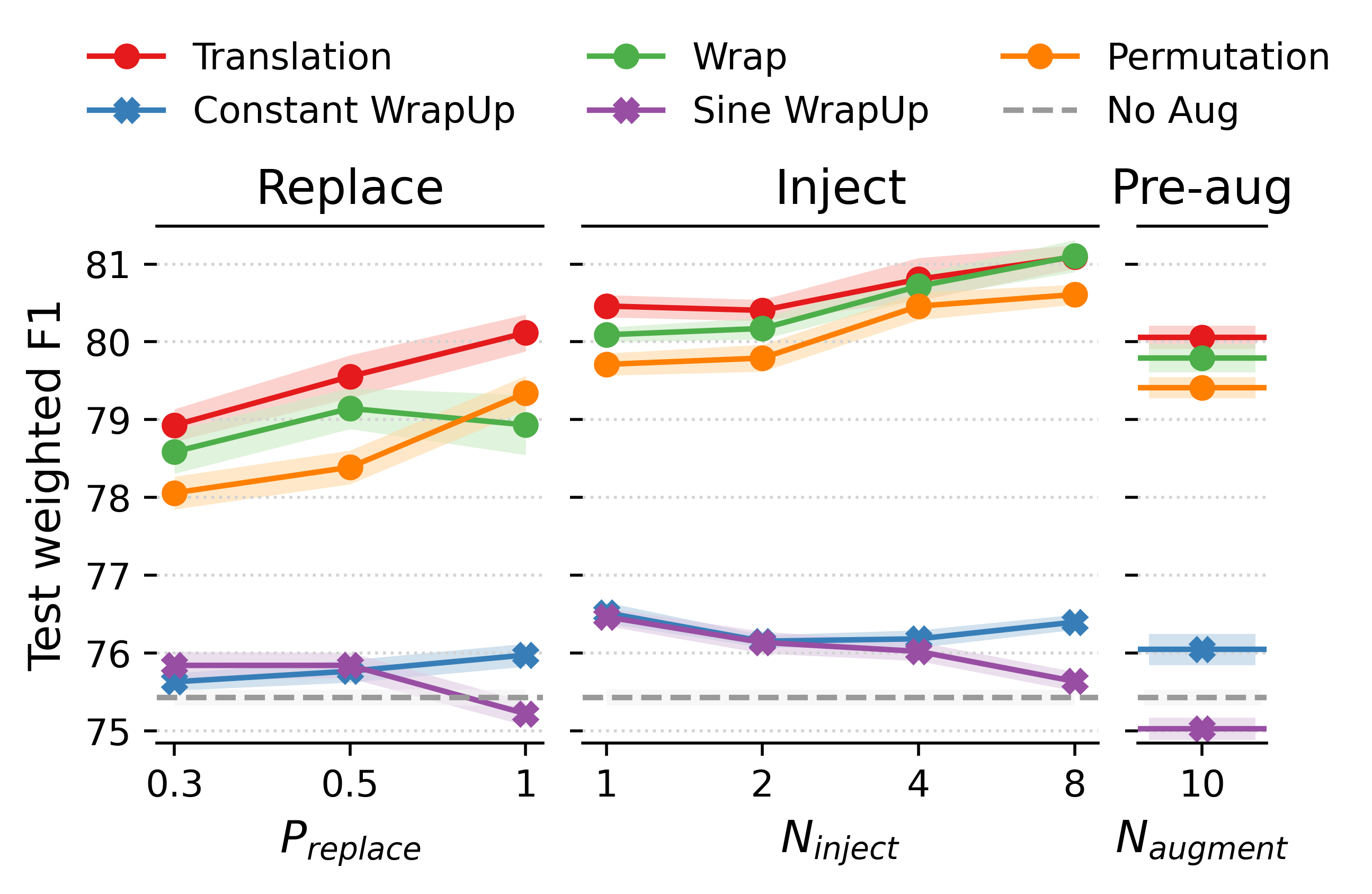} 
    \caption{\MIRAGEA}%
  \end{subfigure}
\vspace{1mm}
   \begin{subfigure}{0.495\textwidth}
    \includegraphics[width=\textwidth]{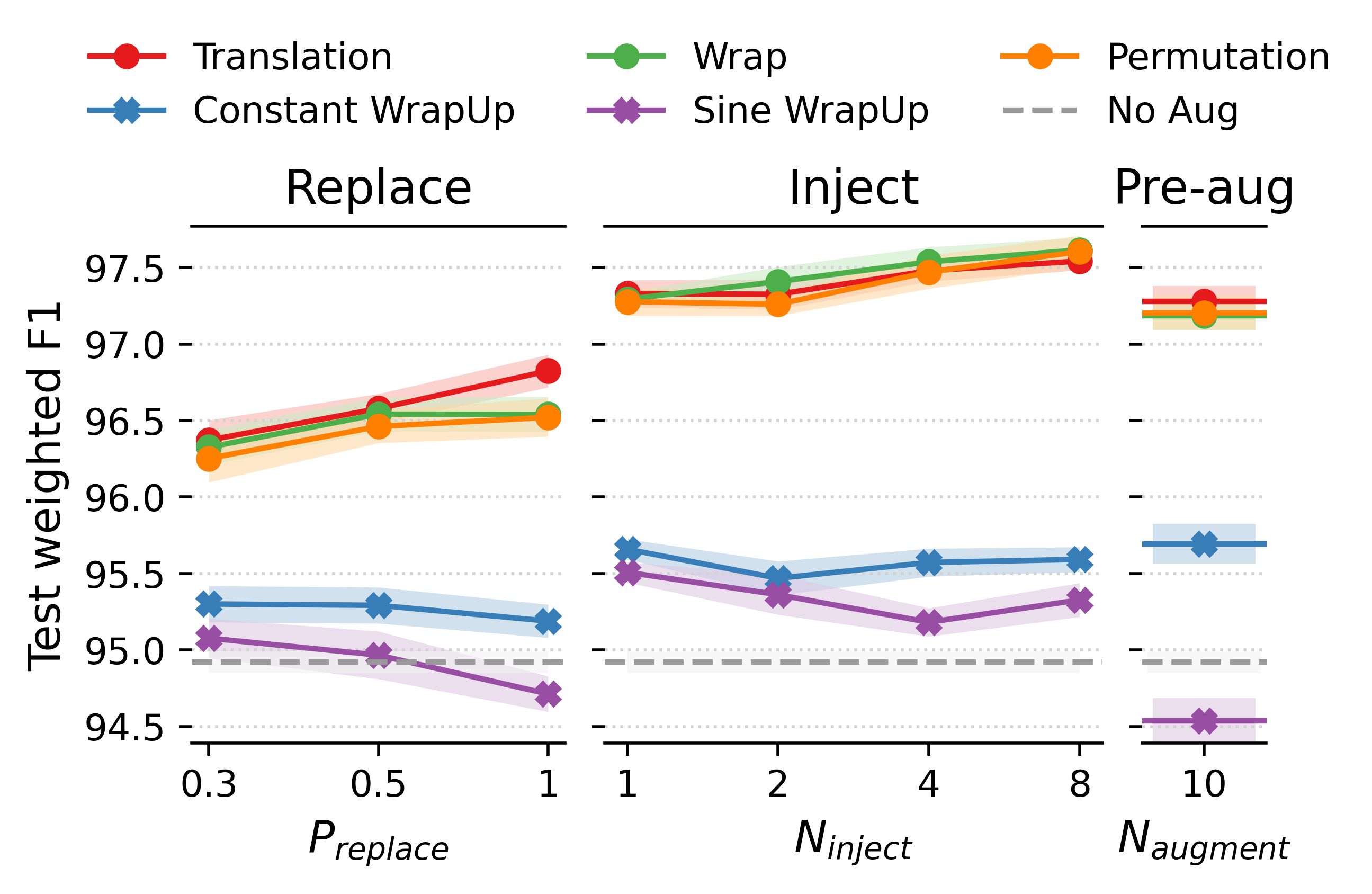} 
    \caption{\MIRAGEB}%
  \end{subfigure}
  
\caption{Comparing \emph{Replace}, \emph{Inject} and \emph{Pre-augment} batch creation policies
\textbf{(G2).}
\label{fig:minibatch-creation-policy}
}
\end{figure}

\vspace{3pt}
\noindent \textbf{Batching policies.}
To show this, we considered the three policies
introduced in Sec.~\ref{sec:method_batching}:
Replace (which randomly substitutes training samples
with augmented ones), 
Inject (which expands batches by adding 
augmented samples), and 
Pre-augment (which expands the whole training set before the training 
start).\footnote{Based on our experience on using
code-bases related to publications, we were unable to pinpoint
if any of those techniques is preferred in CV literature.}
Batching policies are compared against training without
augmentations making sure that each training step has
the same batch size B=1,204.\footnote{For instance, 
when $N_{inject}=1$, a training run
needs to be configured with B=512 as the mini-batches size doubles 
via augmentation.}
Based on Sec.~\ref{sec:results_rank} results, we limited our comparison to 
\{Translation, Wrap, Permutation\}
against \{Sine WrapUp, Constant WrapUp\} 
as representative of good and poor augmentations across the three 
datasets under study. 
We configured Replace with
$P_{replace}\in\{0.3, 0.5, 1\}$,
Inject with $N_{inject}\in\{1,2,4,8\}$
and augmented each training sample 10 times
for Pre-augment.
Fig.~\ref{fig:minibatch-creation-policy} collects
the results with lines showing the average performance while shaded areas
correspond to 95th percentile confidence intervals.
Overall, top-performing augmentations ($\CIRCLE$ marker)
show a positive trend---the higher the volume of augmentations
the better the performance---while poor-performing
augmentations ($\mathbf{\times}$ marker) have small deviations from
the baseline (dashed line). 
Based on performance, we can order 
\emph{Replace} $<$ \emph{Pre-augment} $<$ \emph{Inject},
i.e., the computationally cheaper Pre-augment
is on par with the more expensive Replace when $P_{replace}=1$ 
but Inject is superior to both alternatives.

\noindent\textit{\textbf{Takeaways.}
On the one hand, Inject shows a positive trend
that perhaps continues beyond $N_{inject}>8$.\footnote{The limit
of our experimental campaigns were just bounded by training time and 
servers availability so it is feasible to go beyond the 
considered scenarios.}
On the other hand, the performance gain
may be too little compared to the computational
cost when using many augmentations. For instance,
$N_{inject}=8$ requires 3$\times$
longer training compared to $N_{inject}=1$.
}

\begin{table}[t]
\centering
\caption{Impact of class-weighted sampler on \MIRAGEA \textbf{(G2)}.
\label{tab:g2_weighted_sampling}
}
\small
\begin{tabular}{
rr
@{$\:\:$}r
@{$\:\:$}r
@{$\:\:$}r
@{$\:\:$}r
@{$\:\:$}r
@{$\:\:$}r
@{$\:\:$}r
@{$\:\:$}r
@{$\:\:$}r
}
\toprule
& &
\multicolumn{3}{c}{\bf Majority classes}  & 
\multicolumn{3}{c}{\bf Minority classes}  \\
& 
\bf Cls samp. & 
\multicolumn{1}{c}{\it Pre} &
\multicolumn{1}{c}{\it Rec} &
\multicolumn{1}{c}{\it weight F1} &
\multicolumn{1}{c}{\it Pre} &
\multicolumn{1}{c}{\it Rec} &
\multicolumn{1}{c}{\it weight F1}
\\
\cmidrule(r){2-2}
\cmidrule(r){3-5}
\cmidrule(r){6-8}
\multirow{3}{*}{\bf No Aug}
&  with
&  83.90\extrtiny{.21} &  81.01\extrtiny{.21} &  82.36\extrtiny{.14} 
&  56.63\extrtiny{.38} &  60.78\extrtiny{.26} &  58.18\extrtiny{.21} 
\\
&  without
&  81.60\extrtiny{.23} &  82.93\extrtiny{.19} &  82.16\extrtiny{.12} 
&  62.29\extrtiny{.48} &  58.02\extrtiny{.38} &  59.78\extrtiny{.27}
\\
\cmidrule(r){3-5}
\cmidrule(r){6-8}
&  \it diff
&  \BESTPERF 2.30\extrtiny{.32} &  \WORSTPERF -1.92\extrtiny{.28} & \BESTPERF 0.20\extrtiny{.20} 
&  \WORSTPERF -5.66\extrtiny{.60} &  \BESTPERF 2.76\extrtiny{.46} & \WORSTPERF -1.60\extrtiny{.35}   
\\
\cmidrule(r){2-2}
\cmidrule(r){3-5}
\cmidrule(r){6-8}
\multirow{3}{*}{\bf Translation}
&  with
&  89.12\extrtiny{.09} &  84.26\extrtiny{.11} &   86.43\extrtiny{.08} 
&  60.71\extrtiny{.24} &  68.64\extrtiny{.17} &  63.65\extrtiny{.19} 
\\
&  without
&  85.36\extrtiny{.14} &  86.73\extrtiny{.10} &   85.86\extrtiny{.09} 
&  69.69\extrtiny{.25} &  64.14\extrtiny{.25} &   66.20\extrtiny{.22} 
\\
\cmidrule(r){3-5}
\cmidrule(r){6-8}
&  \it diff
& \BESTPERF 3.77\extrtiny{.06} & \WORSTPERF -2.48\extrtiny{.02} &  \BESTPERF  0.57\extrtiny{.02} 
& \WORSTPERF -8.98\extrtiny{.04} & \BESTPERF 4.50\extrtiny{.09} & \WORSTPERF  -2.55\extrtiny{.05}  
\\
\bottomrule
\end{tabular}
\end{table}

\vspace{3pt}
\noindent \textbf{Class-weighted sampling.}
TC datasets are typically imbalanced (see Table~\ref{tab:datasets}). 
It is then natural to wonder if/how augmentations can help
improve performance for classes with fewer samples, namely \emph{minority classes}.
While the batching policies discussed do not alter
the natural distribution of the number of samples per class,
alternative techniques like Random Over Sampling (ROS) and Random Under Sampling (RUS)
allow to replicate/drop samples for minority/majority classes~\cite{johnson2019survey-class-imbalance}.
A \emph{class-weighted sampler} embodies a more refined
version of those mechanisms and composes training mini-batches 
by selecting samples with a probability inversely proportional 
to the classes size---each training epoch results in a
balanced dataset. When combined with augmentations, this 
further
enhance minority classes variety.

The adoption of a class-weighted sampler seems a good idea
in principle. Yet, the enforced balancing in our experience
leads to conflicting results. We showcase this in Table~\ref{tab:g2_weighted_sampling}
where we show Precision, Recall, and weighted F1 for 20 runs
trained with/without a weighted sampler and with/without Translation
(selected as representative of a good augmentation across datasets).
We break down the performance between majority and minority classes
and report per-metric differences when using or not the weighted sampler.
The table refers to \MIRAGEA but similar results can be obtained
for the other datasets.
Ideally, one would hope to observe only positive differences
with larger benefits for minority classes.
In practice, only the Recall for minority classes improves
and overall we observe a poorer weighted F1 (-0.26 across all classes).
By investigating misclassifications, we found that
majority classes are more confused with 
minority classes and when introducing augmentations those effects are 
further magnified.

\noindent\textit{\textbf{Takeaways.}
Paying too much attention to
minority classes can perturb the overall classifier balance,
so we discourage the use of class-weighted samplers.
}

\begin{figure}[t]
\centering
\begin{subfigure}{0.495\textwidth}
\includegraphics[width=\textwidth]{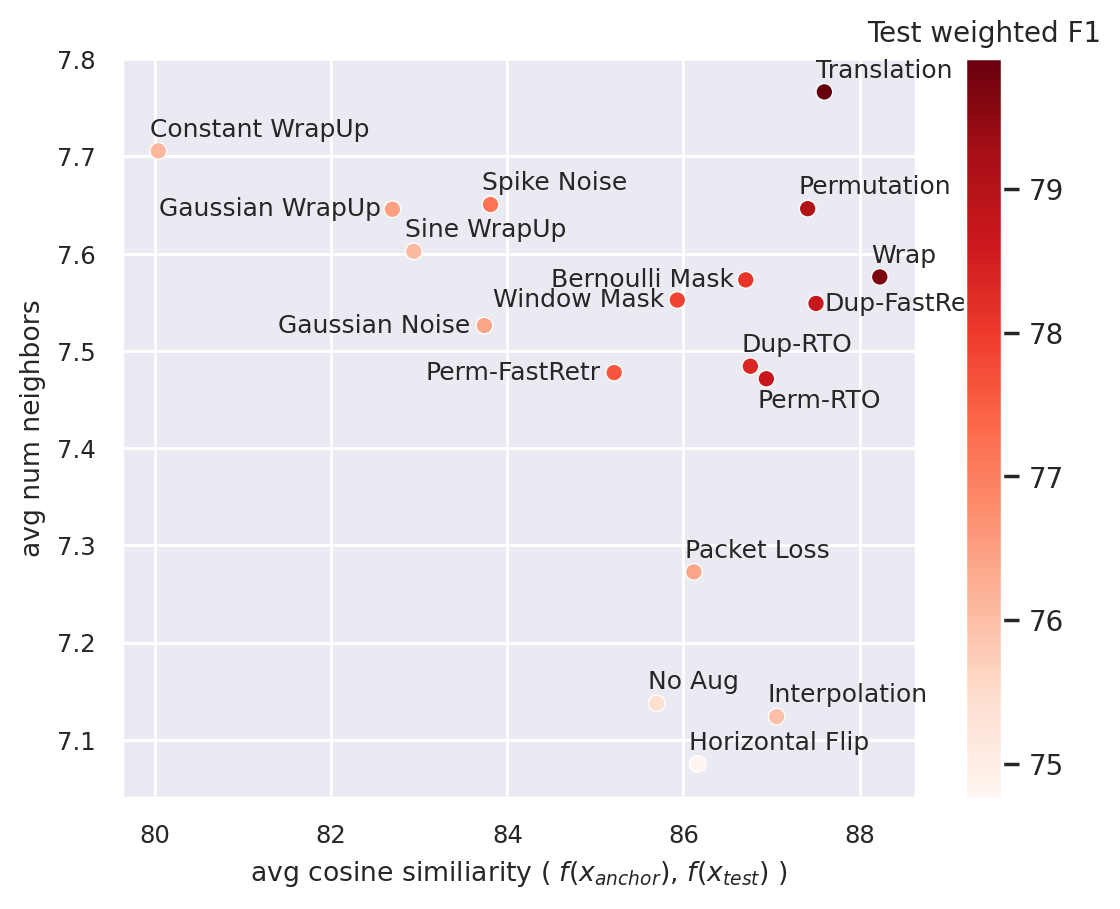}
\caption{\MIRAGEA (anchor=orig $\cup$ aug)}
\end{subfigure}
\begin{subfigure}{0.495\textwidth}
\includegraphics[width=\textwidth]{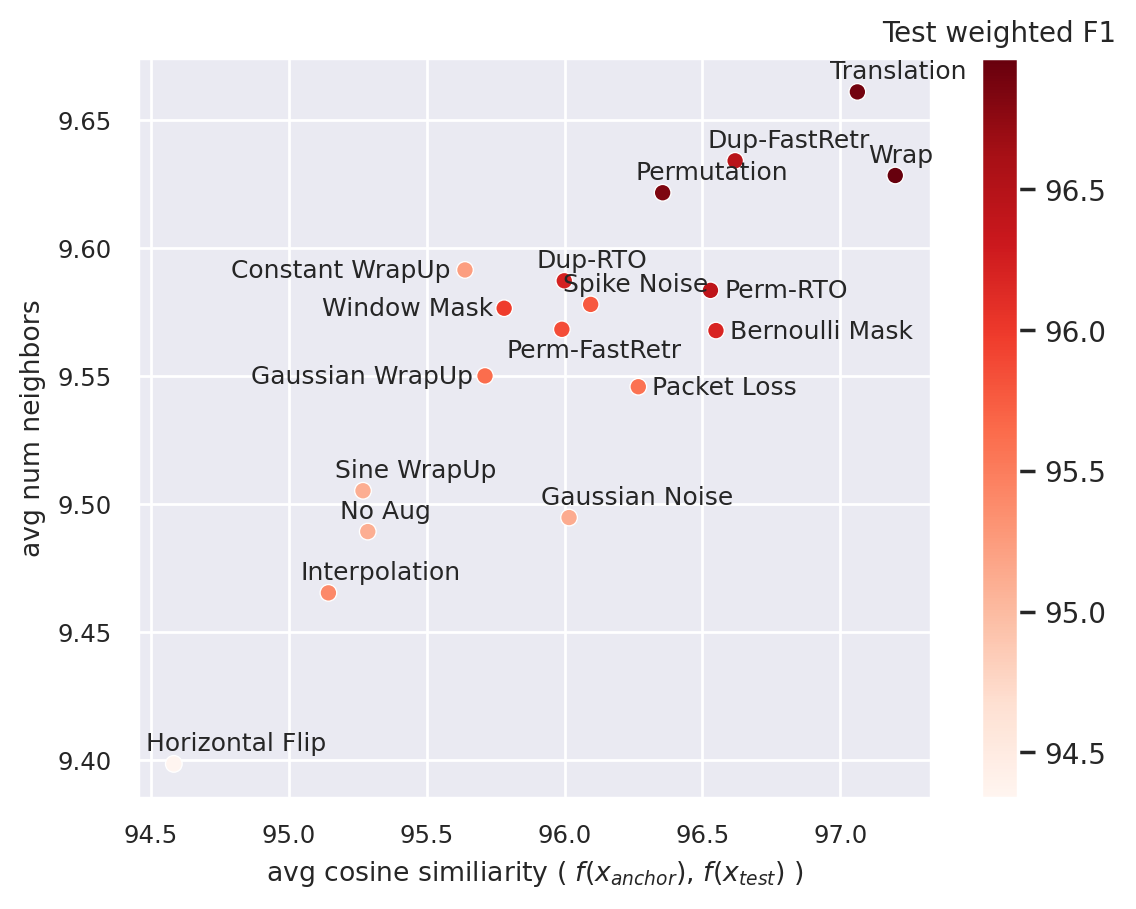}
\caption{\MIRAGEB (anchor=orig $\cup$ aug)}
\end{subfigure}
\begin{subfigure}{0.495\textwidth}
\includegraphics[width=\textwidth]{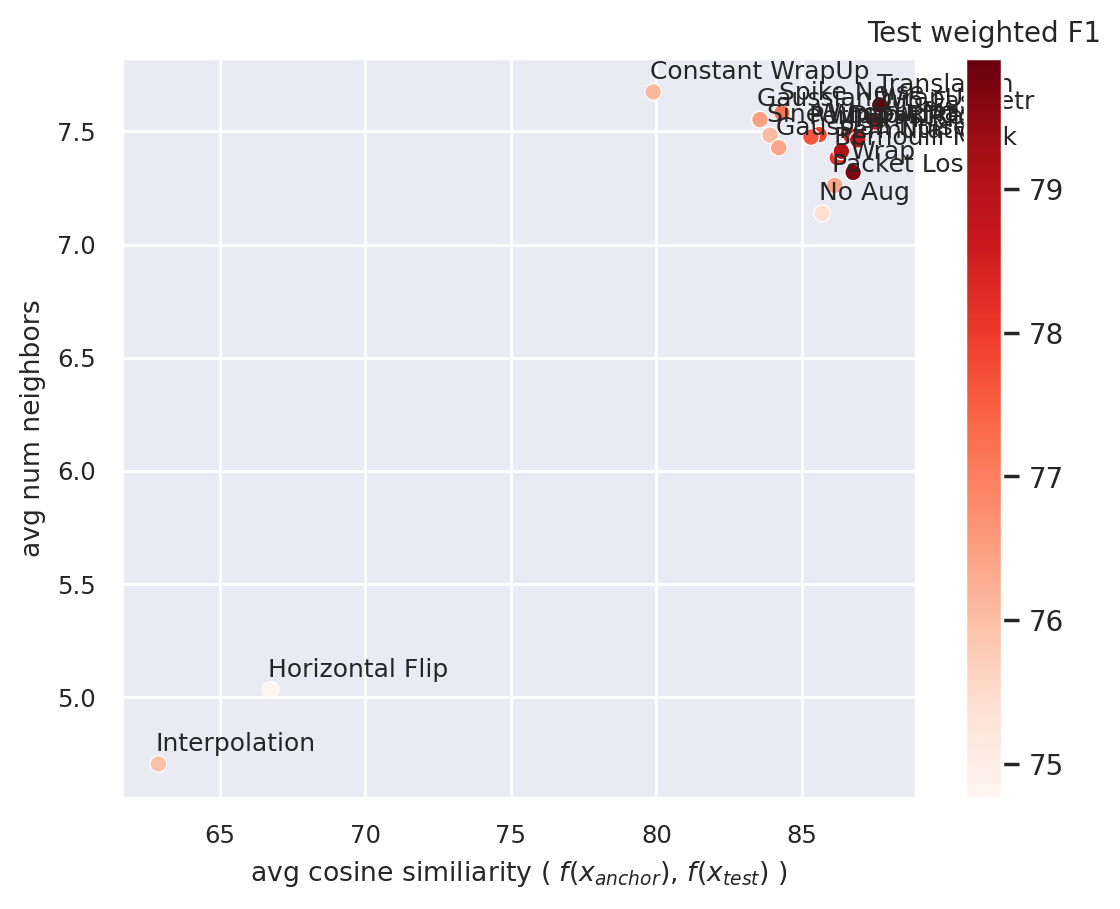}
\caption{\MIRAGEA (anchor=aug)}
\end{subfigure}
\begin{subfigure}{0.495\textwidth}
\includegraphics[width=\textwidth]{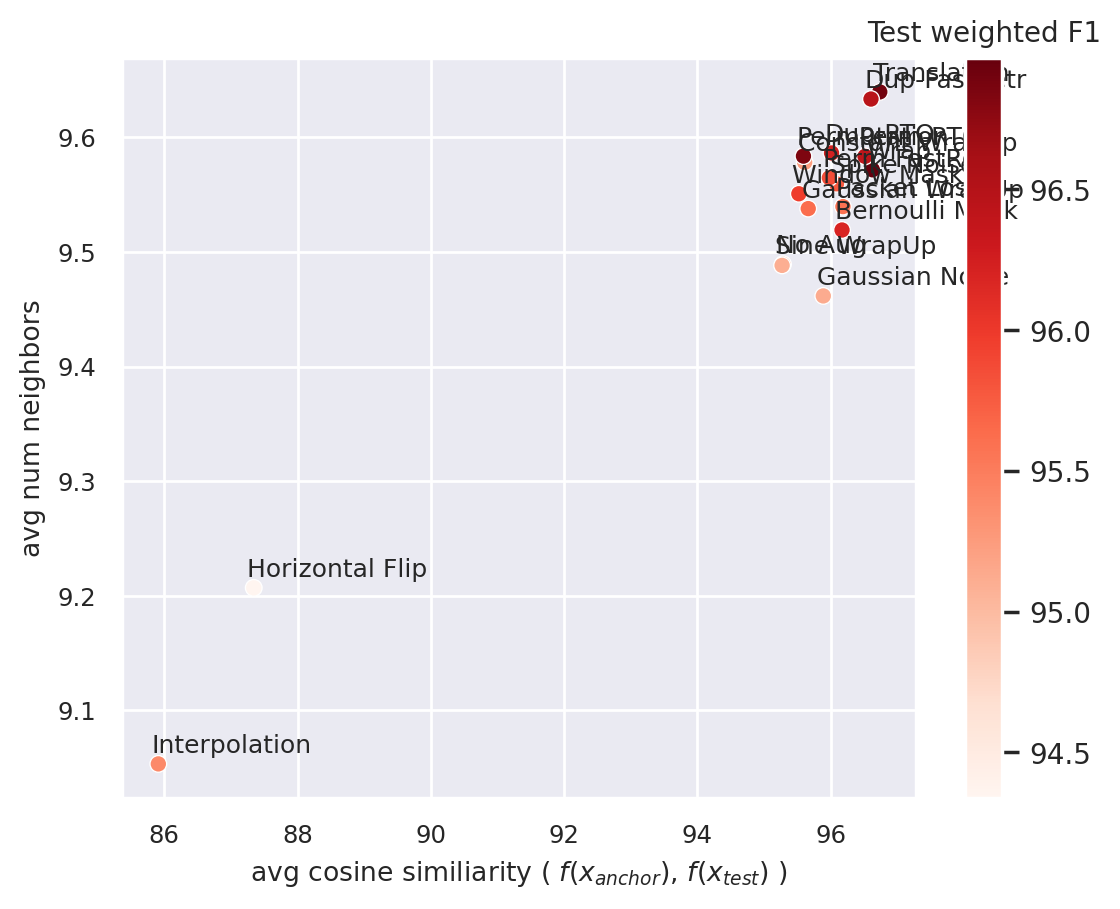}
\caption{\MIRAGEB (anchor=aug)}
\end{subfigure}
\caption{Investigating train, augmented and test samples relationships \textbf{(G3)}.
\label{fig:feature_space_a}
}
\end{figure}

\subsection{Latent space geometry (G3)
\label{sec:results_latent_space}
}

Table~\ref{tab:g1_benchmark} allows to
identify effective augmentations bringing significant benefits 
in terms of model performance.
However, to understand the causes behind
the performance gaps we need to investigate
how original, augmented, and test samples
relate to each other.

\vspace{3pt}
\noindent \textbf{Augmented-vs-Test samples.}
We start our analysis by taking the point of view of the
test samples. Specifically, we investigated
which type of points are found in the 
``neighbourhood'' of a test sample.
To do so, we started
creating ``true anchors'' by projecting
both the original training data and 
5 augmentations of each training
sample---these anchors are ``proxy'' of what
is presented to the model during training. 
Then we projected the test samples and
looked for the closest 10 anchors (based
on cosine similarity) of each
test sample. Finally, we counted how many 
of the 10 anchors share the same label
as the test samples. Results for each
augmentation are reported 
in Fig.~\ref{fig:feature_space_a} for \MIRAGEA and \MIRAGEB 
(similar results holds for \HOMECAMPUS) 
as a scatter plot 
where the coordinates of each point
correspond to the average
number of anchors with the correct label found
and their average cosine similarity with respect to the 
test sample. Each augmentation is color-coded
with respect to its weighted F1 score.

\begin{figure}[!t]
\centering
\begin{minipage}{.69\linewidth}
\begin{subfigure}{0.48\textwidth}
    \includegraphics[width=\textwidth]{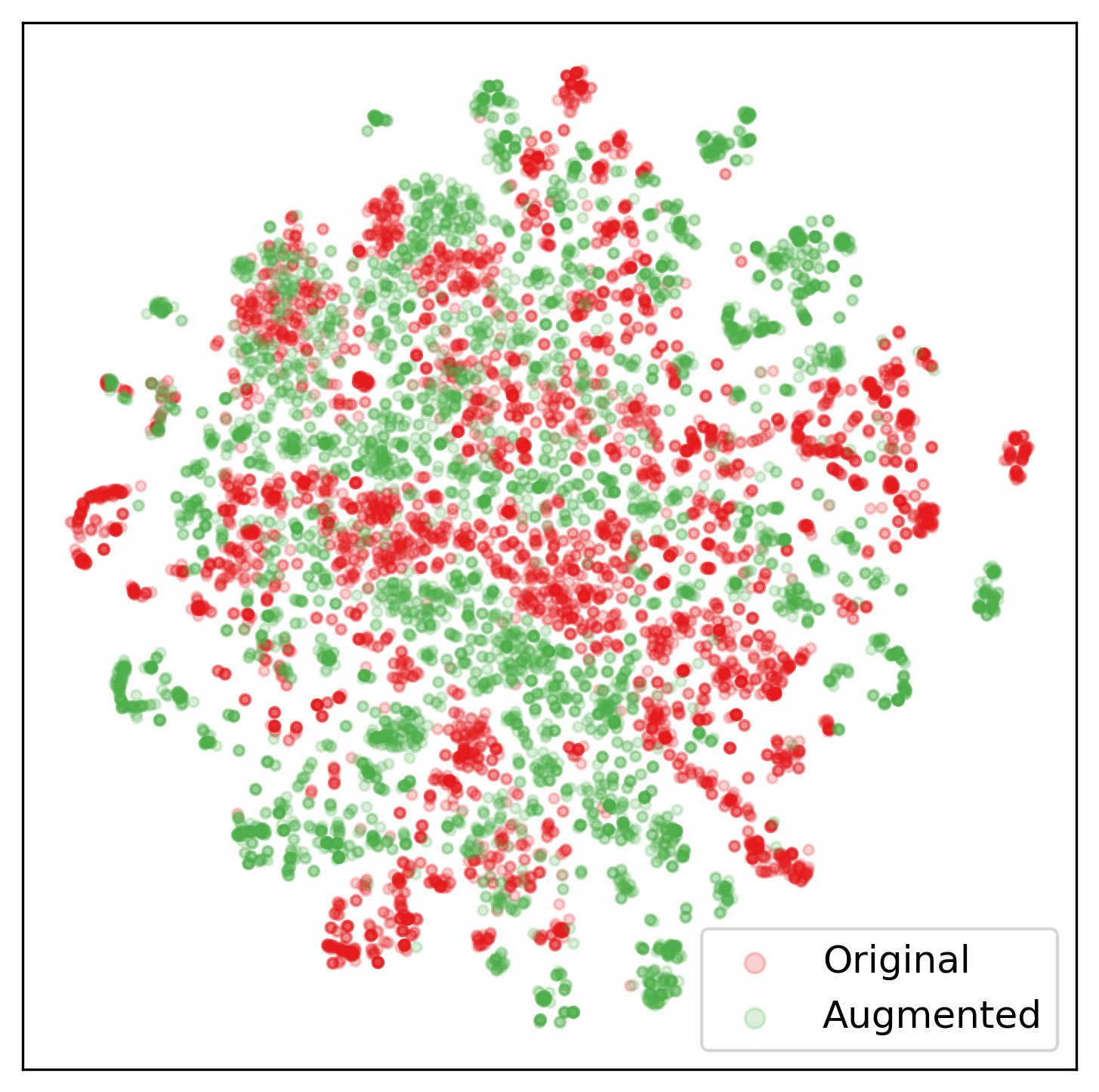}
    \caption{Horizontal Flip.}
    \label{fig:Horizontal Flip_org_vs_aug}
\end{subfigure}
\hfill
\begin{subfigure}{0.48\textwidth}
    \includegraphics[width=\textwidth]{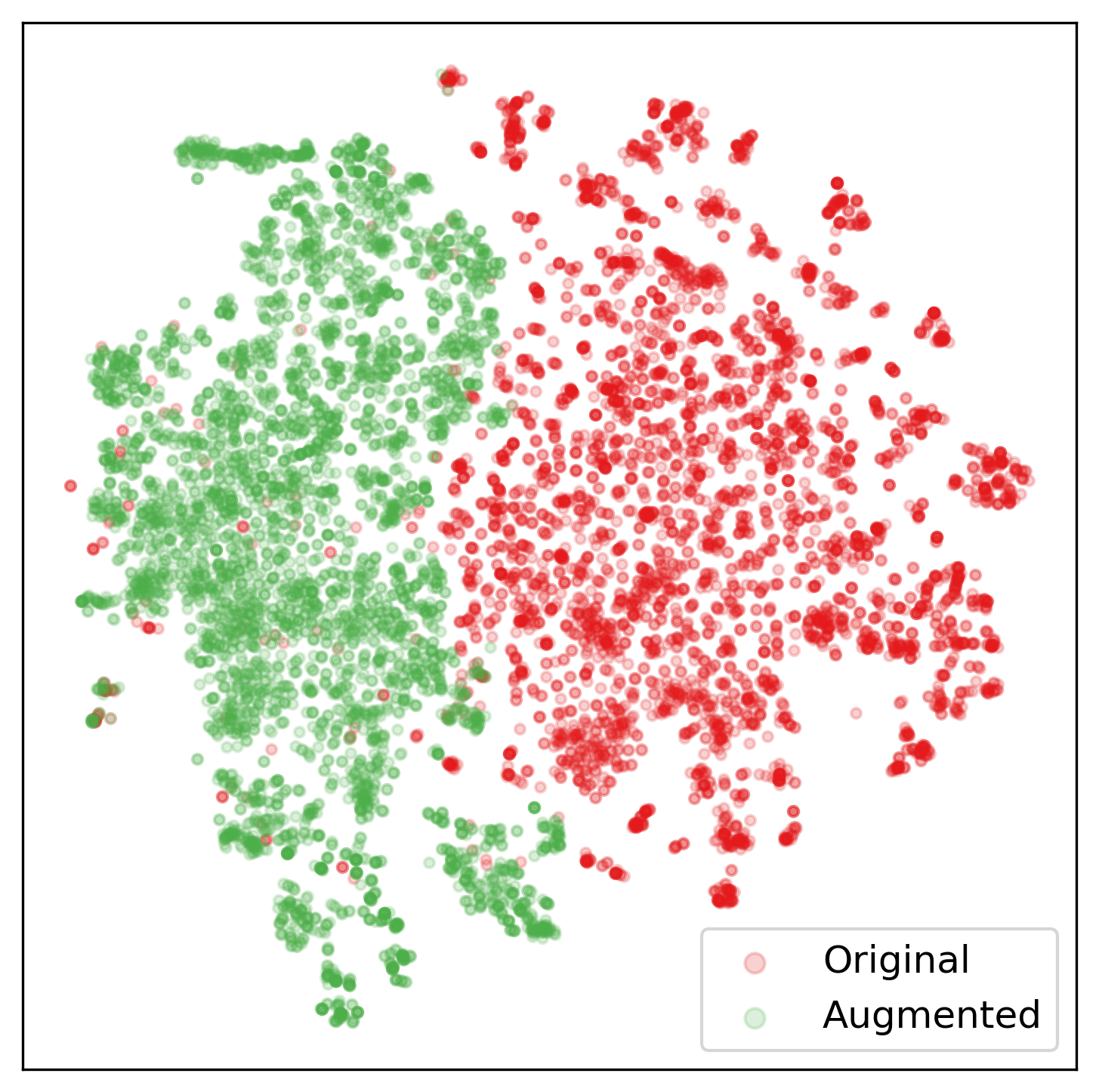}
    \caption{Interpolation.}
    \label{fig:Interpolation_org_vs_aug}
\end{subfigure}
\begin{subfigure}{0.48\textwidth}
    \includegraphics[width=\textwidth]{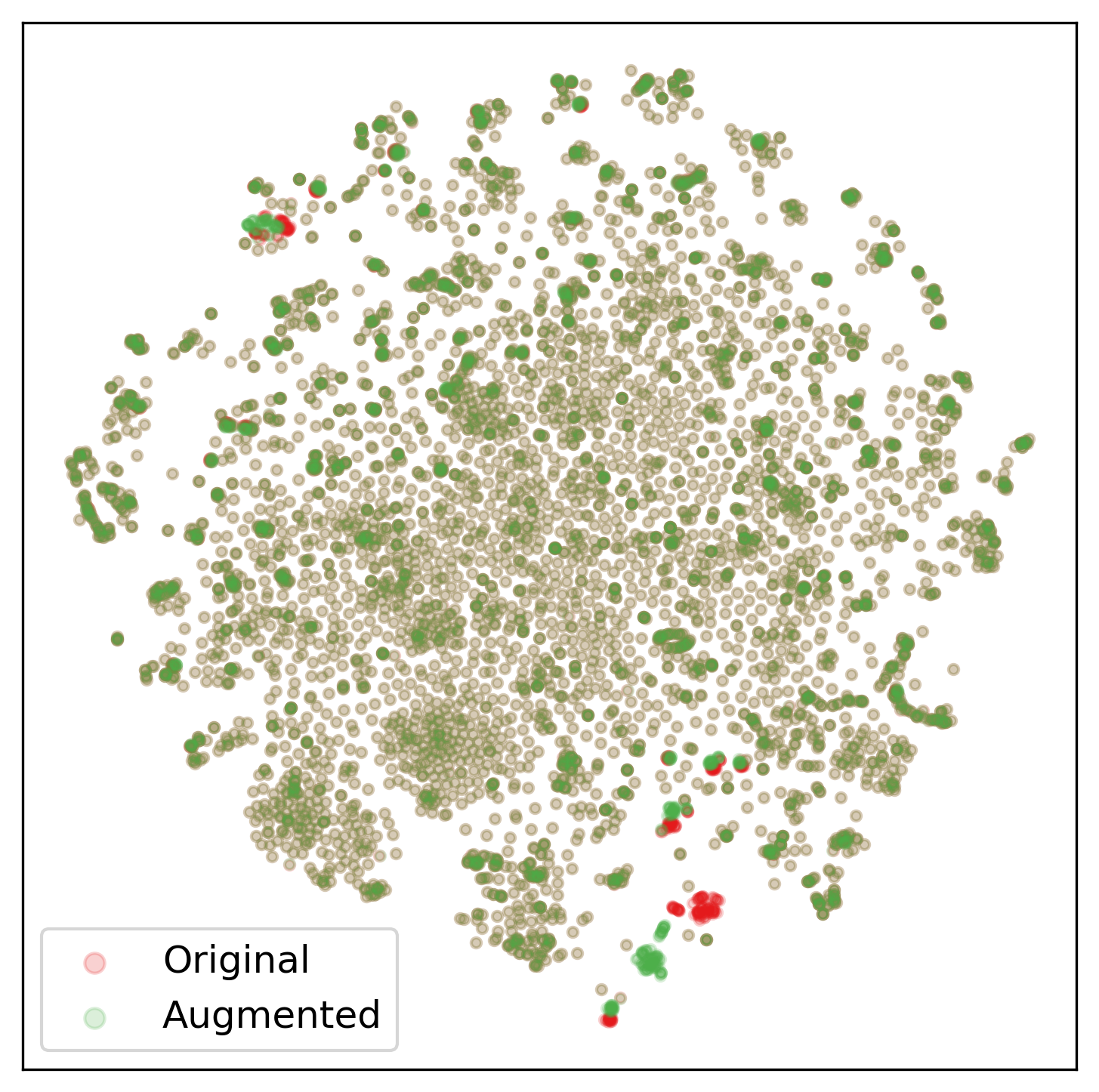}
    \caption{Constant WrapUp.}
    \label{fig:Constant WrapUp_org_vs_aug}
\end{subfigure}
\hfill
\begin{subfigure}{0.48\textwidth}
    \includegraphics[width=\textwidth]{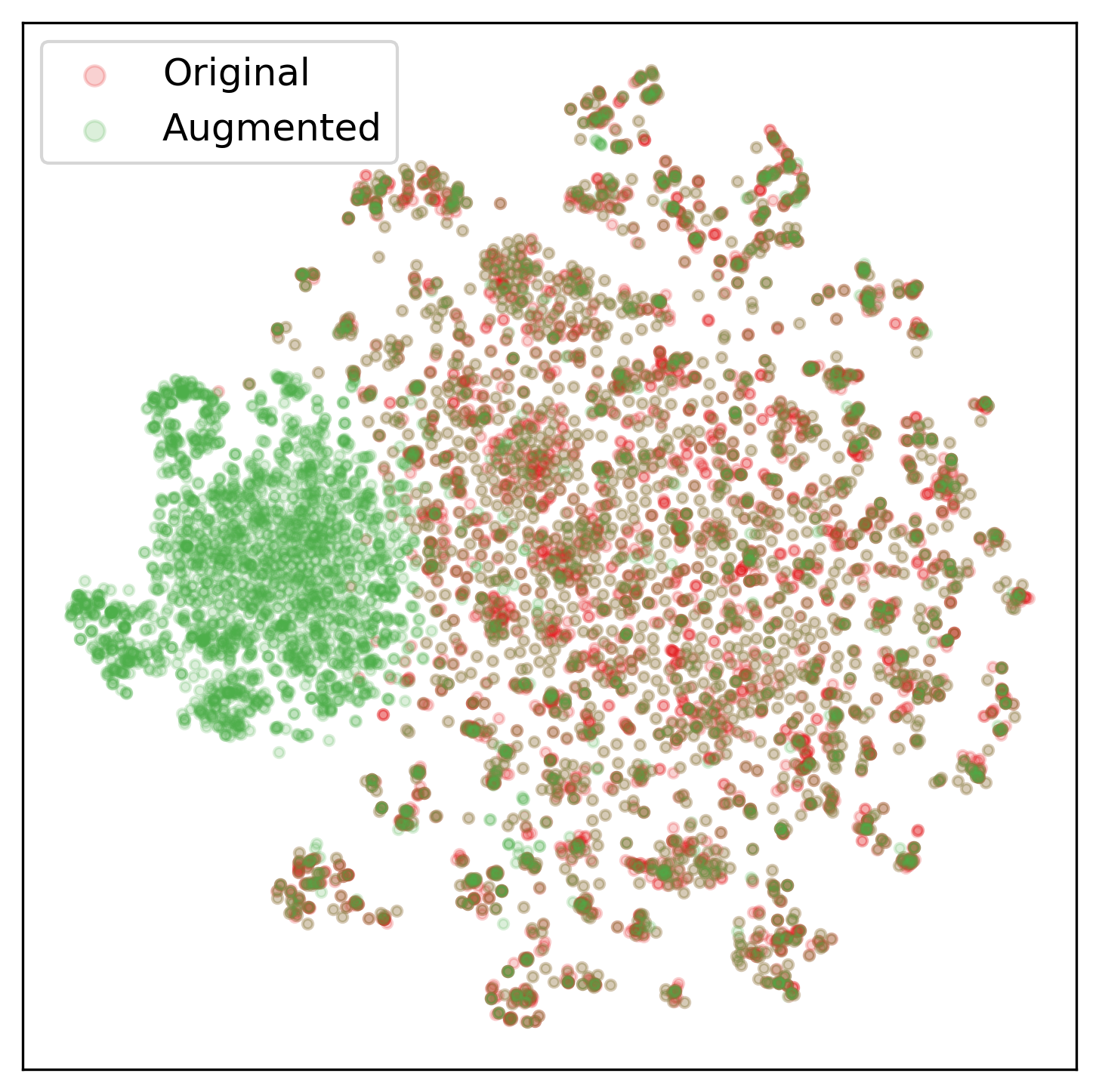}
    \caption{Sine WrapUp.}
    \label{fig:Sine WrapUp_org_vs_aug}
\end{subfigure}
\begin{subfigure}{0.48\textwidth}
    \includegraphics[width=\textwidth]{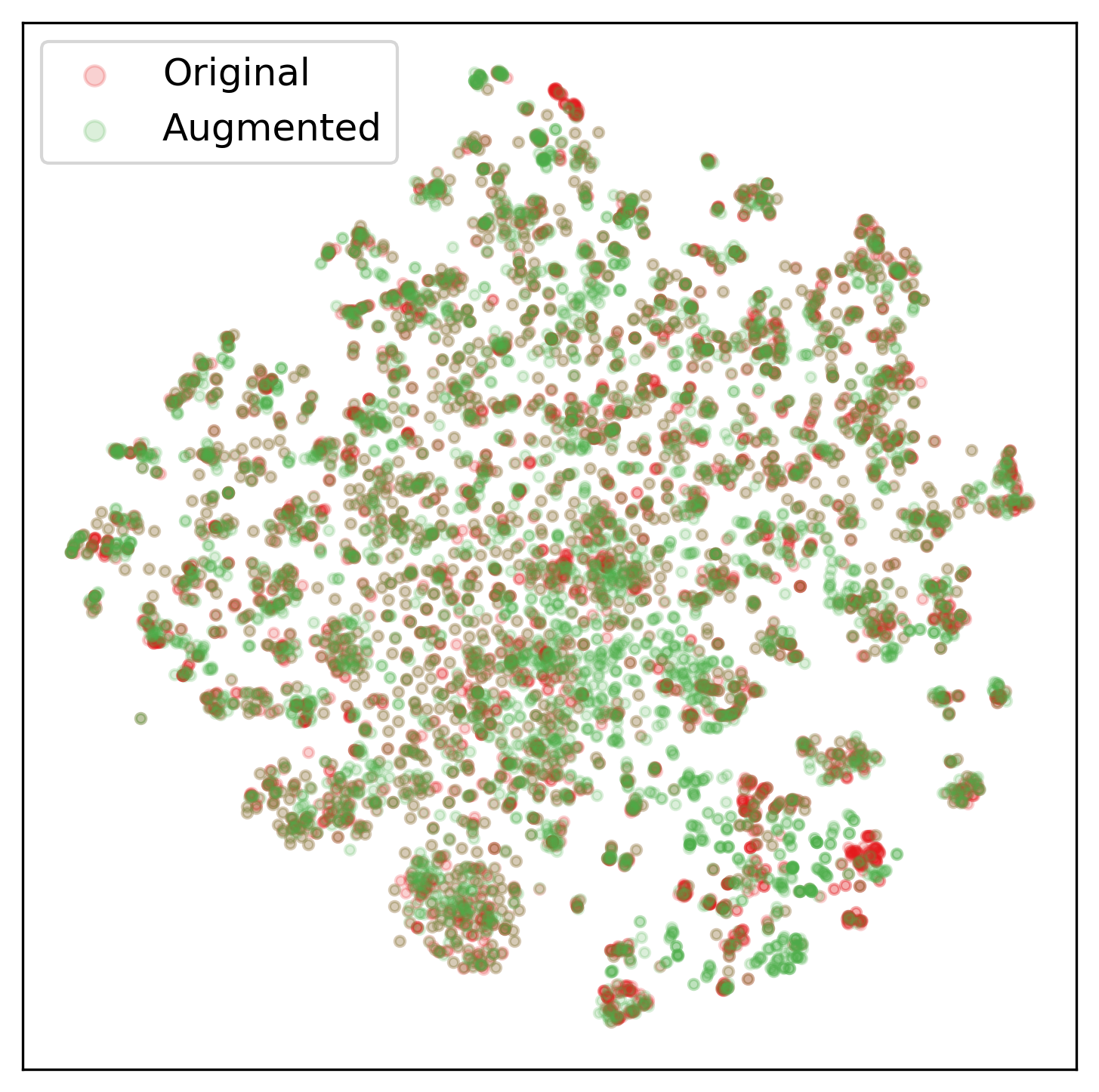}
    \caption{Translation.}
    \label{fig:Translation_org_vs_aug}
\end{subfigure}
\hfill
\begin{subfigure}{0.48\textwidth}
    \includegraphics[width=\textwidth]{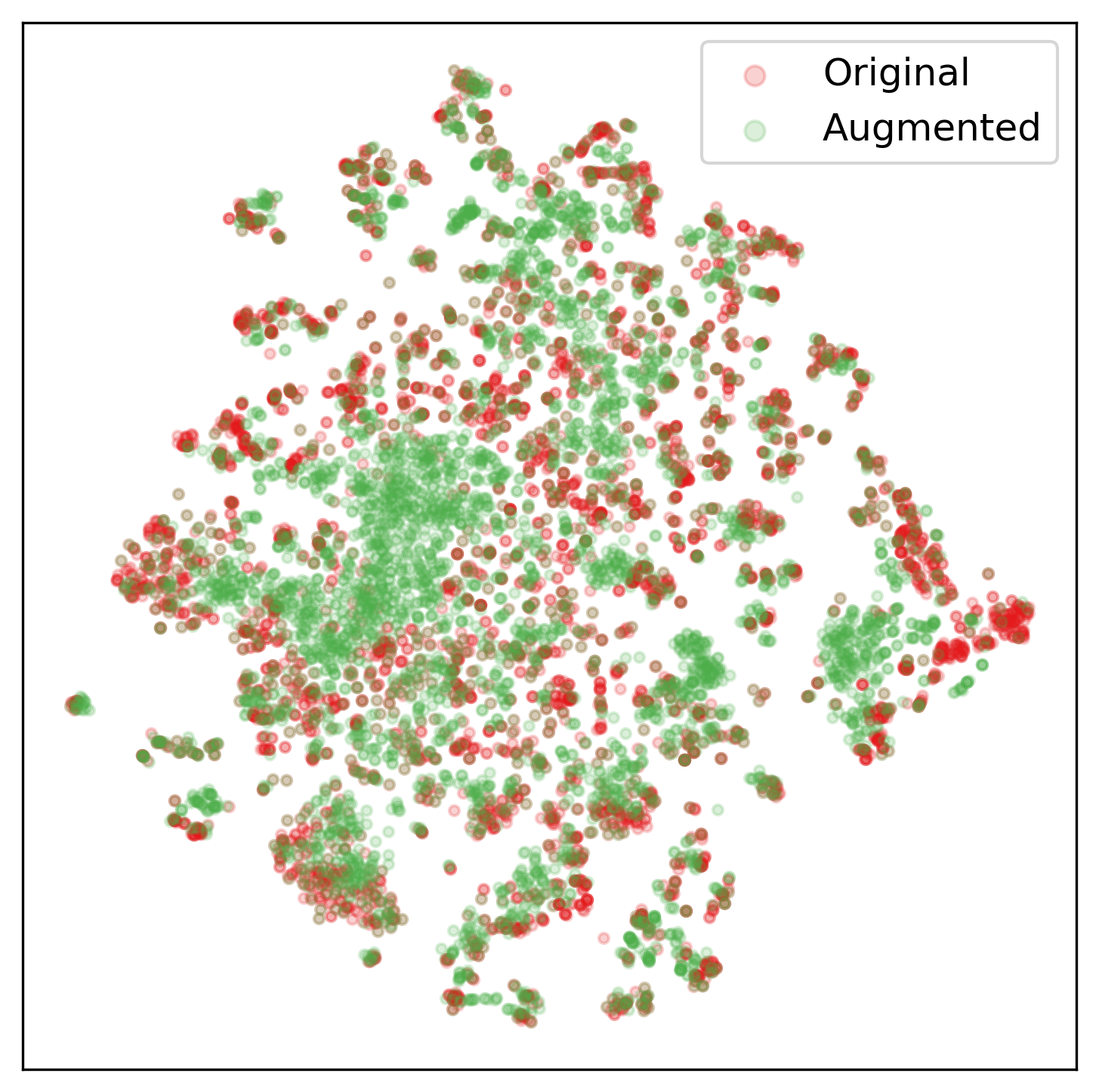}
    \caption{Wrap.}
    \label{fig:Wrap_org_vs_aug}
\end{subfigure}
\end{minipage}
\begin{minipage}{.3\linewidth}
    \vspace{6.5em}
    \begin{subfigure}{\textwidth}
        \includegraphics[width=\textwidth]{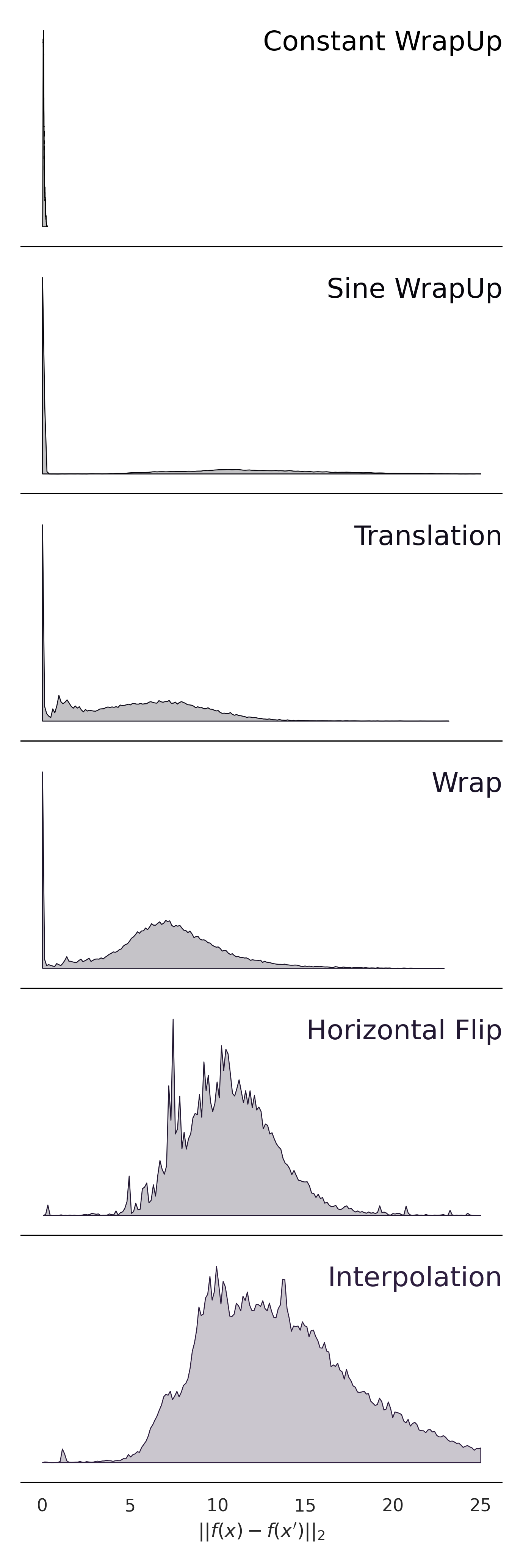}
        \caption{Euclidean dist. KDE.}
        \label{fig:feat_train_dist}
    \end{subfigure}
\end{minipage}

\caption{Comparing original and augmented samples in the latent space \textbf{(G3)}.
\label{fig:latent_space_tsne}
}
\label{fig:figures}
\end{figure}

Despite both metrics vary in a subtle range,
such variations suffice to capture multiple effects.
First of all, considering the layout of the
scatter plot, we expected good transformations
to be placed in the top-right corner.
This is indeed the case as presented 
in Fig.~\ref{fig:feature_space_a} (a-b)
where darker colors (higher weighted F1) concentrate
in the top-right corner. However, 
while \MIRAGEB (Fig.~\ref{fig:feature_space_a}(a))
shows a linear correlation between the two metrics,
\MIRAGEA (Fig.~\ref{fig:feature_space_a}(a)) shows
outliers, most notably Horizontal Flip, Interpolation, and Constant Wrapup.

Fig.~\ref{fig:feature_space_a} (c-d) complement
the analysis by showing results 
when considering only augmented samples as
anchors. Differently from before, now
Horizontal Flip and Interpolation are
found to be the most dissimilar to the test 
samples---this is signaling that augmentations
are possibly disrupting class semantics, i.e.,
they are introducing unnecessary high variety.

Last, for each test sample we looked at the closest 
augmented anchor and the closest original sample anchor
with the same label. The average ratio of those pairwise
distances is centered around 1---augmented samples
``mimic'' test samples as much as the original samples do.

\noindent\textit{\textbf{Takeaways.}
Top-performing augmentations do 
not better mimic test samples compared
to original samples. 
Rather, they help training the feature
extractor $f(\cdot)$ so that projected test samples
are found in neighborhood of points likely to have
the expected label.
}

\vspace{3pt}
\noindent \textbf{Augmented-vs-Original samples.}
We complement the previous analysis by
investigating original $\bf x$ and augmented $\bf x'$ samples relationships.
Differently from before, for this analysis original samples are augmented once.
Then all points are projected in the latent space 
$f({\bf x})$ and $f({\bf x'})$ and visualized
by means of a 2d t-SNE projection.\footnote{
Our model architecture uses a latent space of 256 dimensions
(see Listing~\ref{lst:net})
which the t-SNE representation compresses into a 2d space.
}
We also compute the Kernel Density Estimation (KDE) of the Euclidean distance 
across all pairs. Figure~\ref{fig:latent_space_tsne} 
presents the results for 2 top-performing
(Translate, Wrap) and 4 poor-performing (Constant Wrapup, 
Interpolation, Sine Wrapup, Horizontal Flip) augmentations
for \MIRAGEA. Points in the t-SNE charts
are plotted with alpha transparency, hence
color saturation highlights prevalence
of either augmented or original samples.

Linking back to the previous observations about
Horizontal Flip and Interpolation,
results now show the more
``aggressive'' nature of 
Interpolation---the t-SNE chart is
split vertically with the left (right)
side occupied by augmented (original) samples only
and the Euclidean distance KDEs
show heavier tails.
By recalling their definition, 
while it might be easy to realize
why Horizontal Flip is a poor choice---a client
will never observe the end of the flow before
seeing the beginning, hence they are too artificious---it is difficult to
assess a priori the effect of Interpolation.
Overall, both augmentations
break class semantics.

At the opposite side of the performance range we
find augmentations like
Sine WrapUp and Constant WrapUp.
From Fig.~\ref{fig:latent_space_tsne} we can see that both
introduce little-to-no variety---the Euclidean
distance distributions are centered around zero.
That said, comparing their t-SNE charts we can still observe a major difference
between the two transformations which relates
to their design. Specifically, Constant WrapUp is 
applied only to IAT and introduces negligible modifications
to the original samples.
Conversely, Sine WrapUp is applied on either packet size or IAT.
As for Constant WrapUP, the changes to IAT are subtle,
while variations of packet size lead to generating an extra ``mode''
(notice the saturated cluster of points on the left
side of the t-SNE plot). In other words,
besides the design of the augmentation itself,
identifying a good parametrization is very challenging
and in this case is also feature-dependent.

Compared to the previous, Translate and 
Wrap have an in-between behavior---the 
body of the KDEs
show distances neither too far nor too close
and the t-SNE charts show a non-perfect overlap
with respect to the original samples.
Overall, both these augmentations show positive
signs of good sample variety.

\begin{table}[t]
\centering
\caption{Combining augmentations \textbf{(G4)}.}
\label{tab:g5_ensemble}
\fontsize{9}{9}\selectfont
\begin{tabular}{
    l
    @{$\:\:\:\:\:\:\:\:$}
    l
    @{$\:\:\:\:\:\:\:\:$}
    r
    @{$\:\:\:\:\:\:\:\:\:\:\:\:\:\:\:\:$}
    r
    @{$\:\:\:\:\:\:\:\:$}
}
\toprule
& \bf Augmentation & \MIRAGEA & \MIRAGEB \\
\cmidrule(r){1-2}
\cmidrule(r){3-4}
\bf Baseline
& No Aug   & 75.43\tinytiny{.10}  & 94.92\tinytiny{.07}  \\
\cmidrule(r){1-4}
\multirow{3}{*}{\bf Single}
&  Translation &  4.40\tinytiny{.13} & \WORSTPERF 2.02\tinytiny{.09}  \\
&  Wrap &  \WORSTPERF 4.11\tinytiny{.13} & 2.09\tinytiny{.08}  \\
&  Permutation	& \WORSTPERF 3.67\tinytiny{.13}  & \WORSTPERF 1.97\tinytiny{.09}  \\
\cmidrule(r){1-4}
\multirow{6}{*}{\bf Combined} 
& Ensemble &  4.44\tinytiny{.12}  &   2.18\tinytiny{.09}    \\
&  RandomStack & 4.17\tinytiny{.12}  &  2.18\tinytiny{.09}  \\
&  MaskedStack ($p=0.3$)  &  4.45\tinytiny{.13} &  \BESTPERF 2.26\tinytiny{.09}     \\
&  MaskedStack ($p=0.5$)  & \BESTPERF 4.60\tinytiny{.15}  & \BESTPERF 2.24\tinytiny{.09}     \\
& MaskedStack ($p=0.7$)   & \BESTPERF 4.63\tinytiny{.14}  &  2.18\tinytiny{.10}     \\
\bottomrule
\end{tabular}
\end{table}

\noindent\textit{\textbf{Takeaways.}
Effective transformations operate in a ``sweet spot'':
they neither introduce too little variety---traditional 
policies like Random Over Sampling (ROS) and Random
Under Sampling (RUS)~\cite{johnson2019survey-class-imbalance} 
are ineffective---nor they break classes semantic by 
introducing artificial ``modes''.
}

\subsection{Combining augmentations (G4)
\label{sec:results_ensemble}
}

We conclude our analysis by analyzing the impact of
combining different augmentations. For 
this analysis, we selected 3 top-performing augmentations
and compared their performance when used in isolation
against relying on \emph{Ensemble}, \emph{RandomStack}
and \emph{MaskedStack} (see Sec.~\ref{sec:combining_aug}).
Table~\ref{tab:g5_ensemble} collects results
obtained from 80 modeling runs for each configuration.
Overall, mixing multiple augmentations 
is beneficial but gains are small, i.e., $<$1\%.

\noindent\textit{\textbf{Takeaways.}
While one would expect that mixing good augmentations can only 
improve performance, we note that also CV literature
is split on the subject. If on the one hand combining
augmentations is commonly done in training pipelines,
recent literature shows that such combinations bring
marginal benefits~\cite{muller2021trivialaugment}.
}

\section{Discussion and Conclusions
\label{sec:conclusions}
}

In this work we presented a benchmark of hand-crafted DA 
for TC covering multiple dimensions: a total of 18 augmentations
across 3 families, with 3 policies for introducing 
augmentations during training, investigating the classification
performance sensitivity with respect to augmentations
magnitude and class-weighted sampling across 3
datasets with different sizes and
number of classes. Overall, our results
confirm what previously observed in CV literature---\emph{augmentations are beneficial
even for large datasets, but in absolute terms the gains are dataset-dependent.}
While from a qualitative standpoint, sequence and 
mask augmentations are better suited for TC tasks than amplitude augmentations,
no single augmentation is found superior to alternatives and
combining them (via stacking or
ensembling), even when selecting top-performing ones,
marginally improves performance compared to using augmentations
in isolation.
Last, by investigating the models latent space geometry, we confirm that 
\emph{effective augmentations provide good sample variety}
by creating samples that are neither too similar
nor too different from the original ones which fosters
better data representations extraction
(as suggested by the longer training time).

Despite the multiple dimensions covered,
our work suffers from some limitations.
Most notably,
it would be desirable to include the
larger and more recent \CESNETTLS~\cite{luxemburk2023comnet-cesnettls} and 
\CESNETQUIC~\cite{cesnetquicTMA23} datasets but 
such expansion requires
large computational power.\footnote{
For reference, models trained on \HOMECAMPUS can take up to 6 hours.
Since CESNET datasets contains 100$\times$
the number of samples of \HOMECAMPUS, performing a thorough
exploration of the DA design space is extremely resource demanding.
}
Still related to using large datasets, we can also envision
more experiments tailored to investigate the relationship
between datasets size and augmentations. For instance,
one could sample down a large dataset (e.g., by randomly selecting 1\% or 10\% of the available samples) 
and investigate if augmentations result
more effective with the reduced datasets. In particular,
since Inject shows a positive trend with respect to
its intensity $N_{inject}$ we hypothesize that 
by augmenting a small dataset one can achieve
the same performance as using larger 
datasets---showing these effects are clearly relevant
for TC as collecting and releasing large datasets
is currently a pain point. Last, our campaigns
rely only a CNN-based architecture while assessing
DA with other architectures (e.g., Transformer-based
for time series~\cite{wen2023transformers}) is also relevant.

Ultimately, 
DA modeling campaigns as 
the one we performed require operating
with a grid of configurations
and parameters---it is daunting to explore
the design space by means of
brute forcing all possible scenarios.
While domain knowledge can help
in pruning the search space, it 
can also prevent from considering
valuable alternatives.
For instance, recall that Xie et al.~\cite{Xie2023DACT4TC}
suggest to use augmentations inspired by TCP protocol dynamics.
According to our benchmark, these augmentations are
indeed among the top performing ones, yet not necessarily
the best ones---navigating the search space results in
a balancing act between aiming for qualitative and quantitative results.

We identify two viable options to simplify the design space exploration.
On the one hand, re-engineering
the augmentations so that their parametrization
is discovered during training might 
resolve issues similar to 
what observed for Sine Wrap (see Sec.~\ref{sec:results_latent_space}).
On the other hand,
a more efficient solution would be to rely on 
generative models avoiding the burden of designing
hand-crafted augmentations.
More specifically, we envision
a first exploration based on conditioning the
generative models on the latent space properties learned 
via hand-crafted DA (e.g., the distance between
original and augmented samples should be in the ``sweet spot''). 
Then, we could target the more challenging scenario of
training unconditionally
and verify if effective representations are automatically learned.

Overall, we believe that the performance observed
in our experimental campaigns might still represent a lower
bound and extra performance improvements could be achieved
via generative models. We call for the research community
to join us in our quest for integrating DA and improve TC performance.

\clearpage
\bibliographystyle{splncs04}
\bibliography{bibliography}
\clearpage

\appendix

\begin{figure}[t]
\begin{minipage}{1.01\textwidth}
\centering
\subfloat[\mbox{Overall}]{\includegraphics[height=4cm]{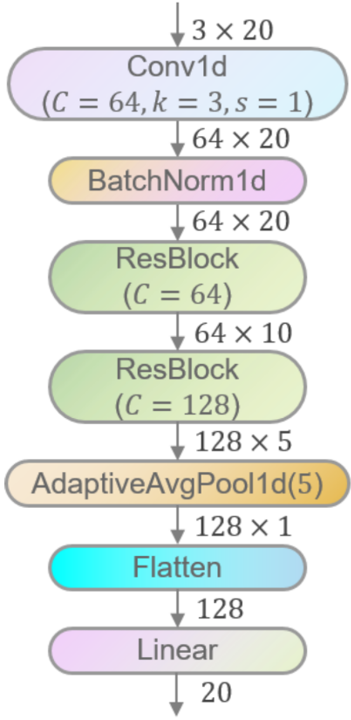}}
\quad\quad\quad
\subfloat[ResBlock]
{\includegraphics[height=4cm]{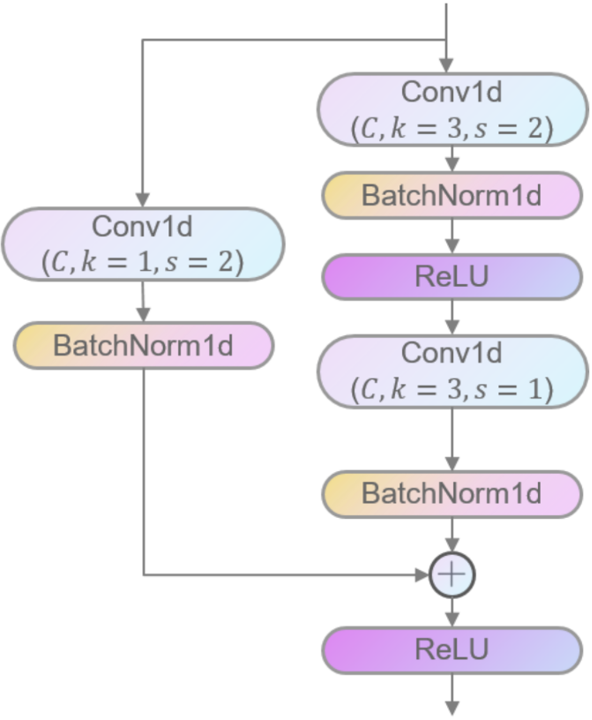}} 
\\
\vspace{5pt}
\scriptsize $C$: number of output channels, $k$: kernel size; $s$: stride
\caption{Model architecture.}
\label{fig:model}
\end{minipage}\hfill
\end{figure}

\begin{centering}
\begin{lstlisting}[caption={Model architecture printout (\MIRAGEA, 20 classes)}, label={lst:net}]
----------------------------------------------------------------
        Layer (type)               Output Shape         Param #
================================================================
            Conv1d-1               [-1, 64, 20]             576
       BatchNorm1d-2               [-1, 64, 20]             128
            Conv1d-3               [-1, 64, 10]          12,288
       BatchNorm1d-4               [-1, 64, 10]             128
            Conv1d-5               [-1, 64, 10]          12,288
       BatchNorm1d-6               [-1, 64, 10]             128
            Conv1d-7               [-1, 64, 10]           4,096
       BatchNorm1d-8               [-1, 64, 10]             128
            Conv1d-9               [-1, 128, 5]          24,576
      BatchNorm1d-10               [-1, 128, 5]             256
           Conv1d-11               [-1, 128, 5]          49,152
      BatchNorm1d-12               [-1, 128, 5]             256
           Conv1d-13               [-1, 128, 5]           8,192
      BatchNorm1d-14               [-1, 128, 5]             256
AdaptiveAvgPool1d-15               [-1, 128, 1]               0
           Linear-16                   [-1, 20]           2,580
================================================================
Total params: 115,028
Trainable params: 115,028
Non-trainable params: 0
----------------------------------------------------------------
Input size (MB): 0.00
Forward/backward pass size (MB): 0.09
Params size (MB): 0.44
Estimated Total Size (MB): 0.53
----------------------------------------------------------------
\end{lstlisting}
\end{centering}

\begin{figure}[H]
\centering
\begin{subfigure}{0.49\textwidth}
\includegraphics[width=\textwidth]{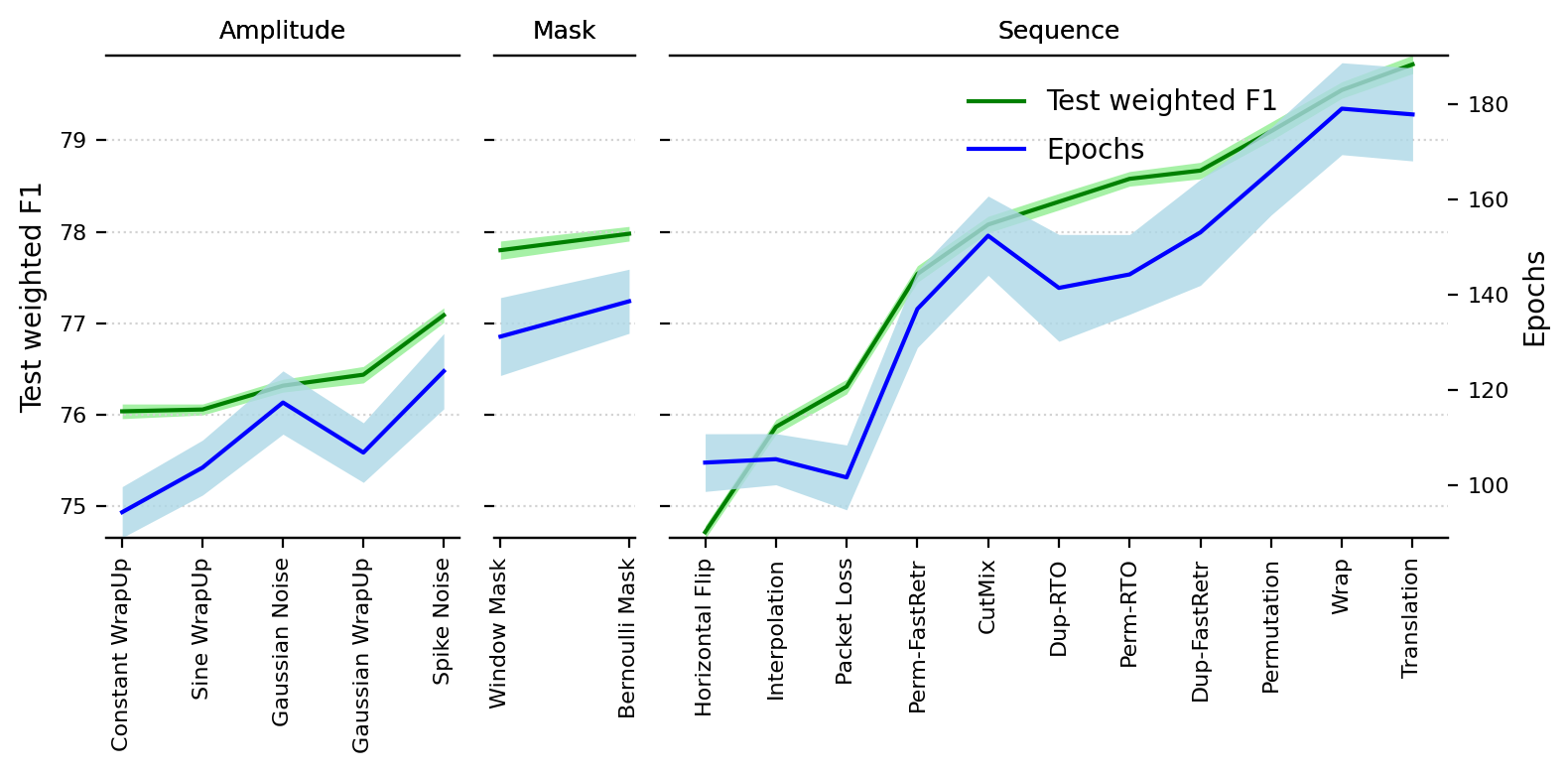}
\caption{\MIRAGEA}
\end{subfigure}
\begin{subfigure}{0.49\textwidth}
\includegraphics[width=\textwidth]{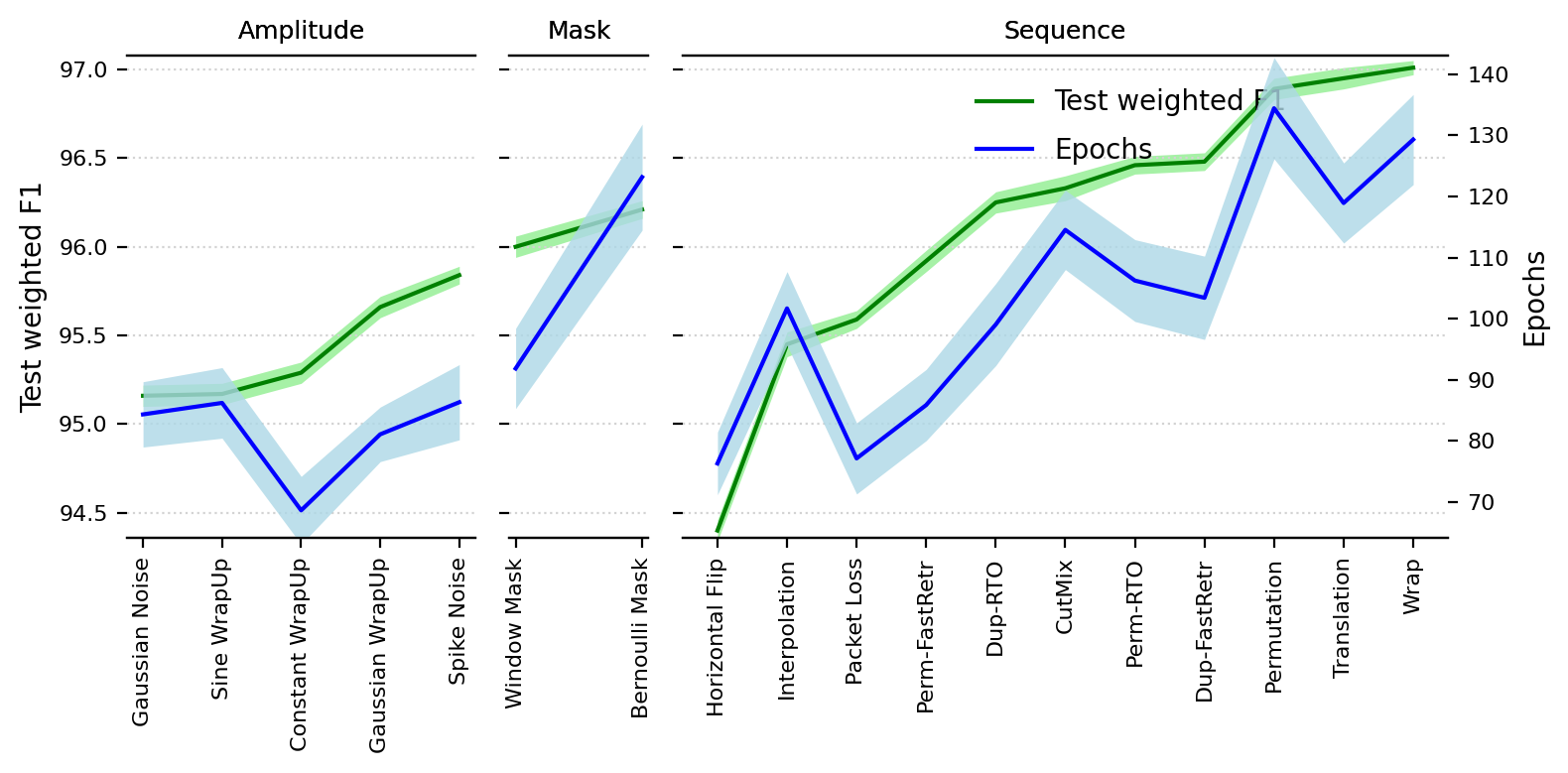}
\caption{\MIRAGEB}
\end{subfigure}
\caption{Comparing performance improvement and training length.
\label{fig:g1_gain_vs_epochs}
}
\end{figure}

\end{document}